\documentclass[conference]{IEEEtran}
\IEEEoverridecommandlockouts

\usepackage{cite}
\usepackage{amsmath,amssymb,amsfonts}
\usepackage{algorithmic}
\usepackage{graphicx}
\usepackage{textcomp}
\usepackage{xcolor}
\def\BibTeX{{\rm B\kern-.05em{\sc i\kern-.025em b}\kern-.08em
    T\kern-.1667em\lower.7ex\hbox{E}\kern-.125emX}}
    
\usepackage{cite}
\usepackage{amsmath,amssymb,amsfonts}
\usepackage{algorithm}[H]
\usepackage{algorithmic}

\usepackage{graphicx}
\usepackage{textcomp}
\usepackage{xcolor}

\usepackage{tcolorbox}
\usepackage{enumitem}
\usepackage{nccmath}
\usepackage{lipsum}
\usepackage{booktabs}
\usepackage{multicol}
\usepackage{multirow} 
\usepackage{makecell}

\usepackage{amsthm}
\newtheorem{theorem}{Theorem}

\newtheorem{lma}{Lemma}

\setlist[itemize]{leftmargin=*}
\setlist[enumerate]{leftmargin=*}
\usepackage{fancyhdr}

\usepackage{hyperref}
\hypersetup{
    colorlinks=true, 
    linktoc=all,     
    linkcolor=blue,  
}
\def\BibTeX{{\rm B\kern-.05em{\sc i\kern-.025em b}\kern-.08em
    T\kern-.1667em\lower.7ex\hbox{E}\kern-.125emX}}

\begin{document}
\newcommand{\SWITCH}[1]{\STATE \textbf{switch} (#1) \textbf{do}}
\newcommand{\ENDSWITCH}{\STATE \textbf{end switch}}
\newcommand{\CASE}[1]{\STATE \textbf{case} #1\textbf{:} \begin{ALC@g}}
\newcommand{\ENDCASE}{\end{ALC@g}}
\newcommand{\CASELINE}[1]{\STATE \textbf{case} #1\textbf{:} }
\newcommand{\DEFAULT}{\STATE \textbf{default:} \begin{ALC@g}}
\newcommand{\ENDDEFAULT}{\end{ALC@g}}
\newcommand{\DEFAULTLINE}[1]{\STATE \textbf{default:} }

\newcommand{\PARALLEL}[1]{\STATE \textbf{parallel for} #1 \textbf{do}}
\newcommand{\ENDPARALLEL}{\STATE \textbf{end parallel}}

\title{Matryoshka: Stealing Functionality of Private ML Data by Hiding Models in Model}

\author{\IEEEauthorblockN{Xudong Pan, Yifan Yan, Shengyao Zhang, Mi Zhang, Min Yang}
\IEEEauthorblockA{School of Computer Science \\
\textit{Fudan University, China}\\
\{xdpan18,yanyf20\}@fudan.edu.cn, shengyaozhang21@m.fudan.edu.cn, \{mi\_zhang,m\_yang\}@fudan.edu.cn}}
\maketitle
\pagestyle{plain}
\lhead{}
\chead{}
\rhead{}
\lfoot{}
\cfoot{}
\rfoot{\thepage}

\begin{abstract}
High-quality private machine learning (ML) data becomes a key competitive factor for AI corporations. Despite being carefully stored in local data centers, a private dataset may still suffer from piracy once a deep neural network (DNN) model trained on the dataset is made public or accessible as a prediction API. If it is the innate tension between the open model interface and the private dataset that leaves an exploitable attack surface, \emph{we wonder if a private dataset with no exposed interface  can be impregnable}.   

In this paper, we present a novel insider attack called \emph{Matryoshka}, which employs an irrelevant scheduled-to-publish DNN model as \textit{a carrier model} for covert transmission of multiple secret models which memorize the functionality of private ML data stored in local data centers. Instead of treating the parameters of the carrier model as bit strings and applying conventional steganography, we devise a novel parameter sharing approach which exploits the learning capacity of the carrier model for information hiding. Matryoshka simultaneously achieves: (i) \textit{High Capacity} --  With almost no utility loss of the carrier model, Matryoshka can hide a $26\times$ larger secret model or $8$ secret models of diverse architectures spanning different application domains in the carrier model, neither of which can be done with existing steganography techniques; (ii) \textit{Decoding Efficiency} -- once downloading the published carrier model, an outside colluder can exclusively decode the hidden models from the carrier model with only several integer secrets and the knowledge of the hidden model architecture; (iii) \textit{Effectiveness} -- Moreover, almost all the recovered models have similar performance as if it were trained independently on the private data; (iv) \textit{Robustness} -- Information redundancy is naturally implemented to achieve resilience against common post-processing techniques on the carrier before its publishing; (v) \textit{Covertness} -- A model inspector with different levels of prior knowledge could hardly differentiate a carrier model from a normal model.
\end{abstract}



\section{Introduction}
``\textit{The availability of large datasets are key enablers of deep
learning}'', as stated by Bengio et al. in the Turing lecture \cite{Bengio2021DeepLF}. High-quality private machine learning (ML) datasets become critical assets for AI corporations to train performant ML models \cite{hai_report}. As the dataset preparation is highly time-consuming and labor-intensive, it is common for relevant parties to hold the private ML data as confidential properties \cite{Wenskay1990IntellectualPP}.

Despite the ML data being carefully curated in local data centers isolated from the open network \cite{google_security}, recent research on ML privacy \cite{Tramr2016StealingML,CorreiaSilva2018CopycatCS,Chandrasekaran2020ExploringCB,Jagielski2020HighAA,Hu2021StealingML} reveals, once a deep neural network (DNN) finishes its training process on a private dataset, the model immediately becomes an exploitable source for privacy breach. By interacting with the trained model in a full-knowledge manner (i.e., with known parameters, model architecture, etc.) or via the prediction API, attacks are known to infer global/individual sensitive attributes of the training data \cite{Fredrikson2015ModelIA,Ganju2018PropertyIA,Melis2019ExploitingUF,Fredrikson2014PrivacyIP}, infer data membership \cite{Shokri2016MembershipIA,Salem2019MLLeaksMA}, or even steal the full functionality of the private datasets from the trained model \cite{Tramr2016StealingML,CorreiaSilva2018CopycatCS,Jagielski2020HighAA}. The success of these previous attacks reflect the tension between the confidential data and the openly accessible trained model. We wonder, if blocking this source of leakage, whether a private dataset with no exposed open interfaces is immune to data stealing attacks.


\noindent\textbf{Our Work.}  We are the first to discover \textit{the severe leakage of data functionality can happen even for private ML datasets with no exposed public interface}. Instead of exploiting a trained model from the outside, our work proposes a novel insider attack which can be conducted by DNN engineers who have access to the target private dataset(s) and is responsible for using an irrelevant dataset to train an independent scheduled-to-publish model (i.e., the \textit{carrier} model). In popular data security models adopted by many AI corporations, potential attackers with such a role is common (\S\ref{sec:data_security_study}). In other words, our work for the first time reveals a new attack surface where an attacker can steal the functionality of private ML data with no contact to any models trained on the victim dataset(s). 

\begin{figure}[t]
\begin{center}
\includegraphics[width=0.5\textwidth]{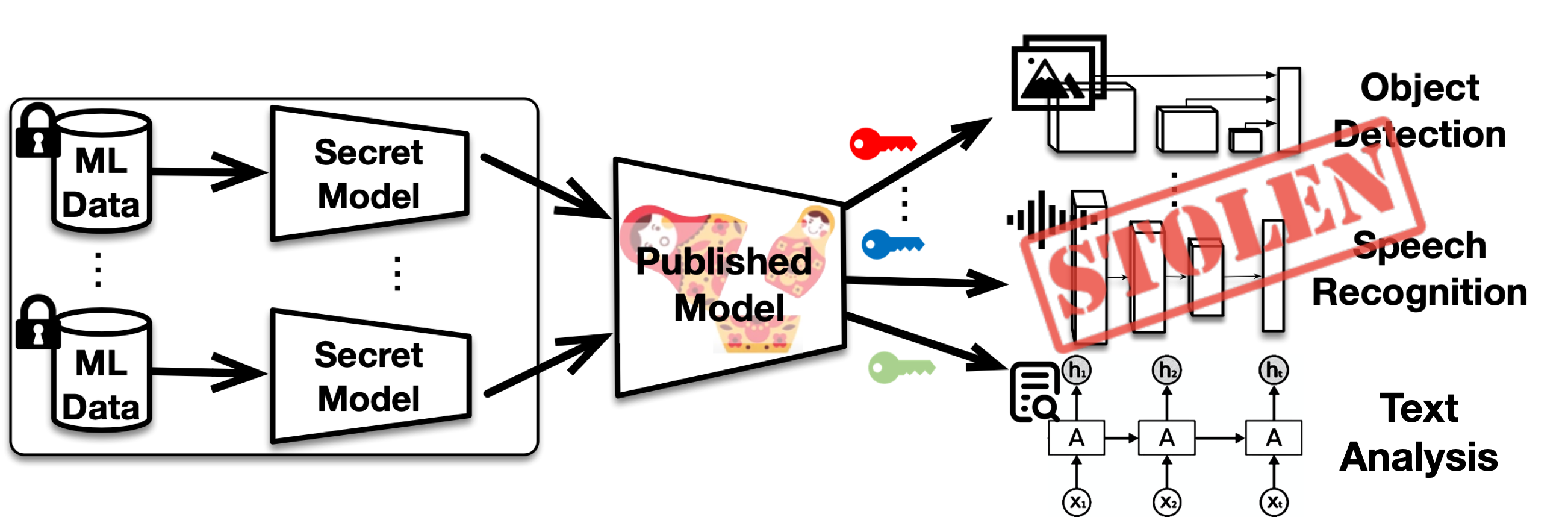}
\caption{The attack scenario of Matryoshka.}
\label{fig:intro}
\end{center}
\vspace{-0.3in}
\end{figure}
As Fig.\ref{fig:intro} shows, to break the privacy of ML data with no exposed interface, our attack employs an irrelevant scheduled-to-publish DNN model as \textit{a carrier model} for covert transmission of multiple task-oriented models trained on different private datasets (i.e., the victims) for functionality stealing. Once an outside colluder decodes these \textit{secret models} from the carrier model, the obtained task-oriented models would immediately share the similar functionality as if trained on the private data. We name this new attack class as \textit{Matryoshka} (nesting dolls in Russian\footnote{By convention, Matryoshka dolls inside (i.e., the secret models) is more
precious than the outer doll which holds them (i.e. the carrier model).}), which literally reflects our novel attack methodology of \textit{hiding models in model}, or \textit{model hiding}.

The key challenge of model hiding is where to find the sufficient encoding capacity for the millions or even billions of floating-point parameters in a secret DNN. Simply treating the parameters of the carrier model as bit strings, existing steganography techniques \cite{Cheddad2010DigitalIS,Yang2019RNNStegaLS,Djebbar2012ComparativeSO,Song2017MachineLM,Uchida2017EmbeddingWI} are strictly upper bound in their encoding capacity, which is far smaller than the size of the carrier model. This bottleneck substantially weakens their flexibility in model hiding. For example, they are unable to encode a secret model in conventional carrier medium (e.g., texts or images), to encode a larger secret model in a small carrier model, or to encode multiple secret models in one carrier model simultaneously for different attack purposes. Moreover, previous studies usually attempt to encode highly heterogeneous data into the parameters of the carrier model. Without special designs, the performance of the carrier model may decrease to an unacceptable level to raise doubts. Finally, the robustness of the existing data hiding techniques may also influence the secret models when the potential obfuscation is conducted on the carrier model \cite{Provos2003HideAS}.

At the core of Matryoshka, we devise a new parameter sharing mechanism which instead exploits the enormous learning capacity of DNN to implement a much larger encoding capacity for model hiding. Intuitively, our parameter sharing mechanism is based on a data structure called \textit{ParamPool}, which stores an array of learnable parameters that are \textit{reusable}, i.e., they are designed to fill into different layers of a given DNN for multiple times, and allow cyclic access. In other words, multiple DNNs, including either the secret models and the carrier model, can be generated from the same ParamPool by recycling the parameters inside  (\S\ref{sec:atk:morph}). To encode the secret models in one carrier model, we jointly train the carrier model along with the secret models with the parameters generated from the ParamPool. During the learning process, the error propagates and accumulates to the corresponding parameter in the ParamPool, where the update finally happens. Our insight behind using ParamPool for parameter sharing comes from a pilot study which observes quite similar parameter distribution of a wide range of well-trained DNNs on various application domains (\S\ref{sec:atk:joint}). After the carrier model is published online, the outside colluder can decode the ParamPool from the carrier model with a small number of secret integer keys, and assemble the secret model(s) for privacy breach (\S\ref{sec:atk:decode}). 

\noindent\textbf{Our Contributions.} As a newly discovered threat to private ML data, our proposed Matryoshka attack implements the following key properties for effective and covert transmission of secret models which steal the data functionality.

\noindent$\bullet$\textbf{ High Capacity}: With almost no utility loss ($< 1\%$), Matryoshka can hide a $26\times$ larger secret model, or $8$ secret models of different architectures spanning image, speech and text applications at one time in the carrier model, \textit{neither} of which is feasible with existing steganography techniques. 

\noindent$\bullet$\textbf{ Decoding Efficiency}: The secret models can be efficiently decoded and assembled from the carrier model with \textit{only several integer secrets} exclusively known to the colluder, provided that the secret models have standard architectures.

\noindent$\bullet$\textbf{ Effectiveness}: Almost all the recovered secret models either have similar performance as when they are independently trained on the private dataset.

\noindent$\bullet$\textbf{ Robustness}: Matryoshka naturally implements information redundancy \cite{Neumann1956ProbabilisticLA} via the usage of a smaller ParamPool, which helps in achieving resilience against possible post-processing techniques on the carrier model before its publishing. 

\noindent$\bullet$\textbf{ Covertness}: Besides, we also provide a preliminary discussion on the covertness of Matryoshka when a detector with different levels of prior knowledge on the attack attempts to differentiate a carrier model from a normal one (\S\ref{sec:evasion}), which we hope may foster future defense studies.

\section{Related Work}
\label{sec:related}
\noindent\textbf{Privacy Attacks on Training Data.}
Recently, trained ML models are proved to be highly exploitable for attackers to steal private information about the training data. One branch of attacks focuses more on the privacy of the training data as a whole. Inferring global sensitive information of the training data, \textit{model inversion attack}  \cite{Fredrikson2014PrivacyIP,Fredrikson2015ModelIA} and \textit{property inference attack} \cite{Ateniese2015HackingSM,Ganju2018PropertyIA} respectively target at revealing the class representatives of a training data (e.g., the average face of an identity when attacking a face recognition model) and whether a first-order predicate holds on the training data (e.g., whether a face dataset contains no whites). Stealing the functionality of a private training dataset, \textit{model extraction attack} constructs a surrogate model by distilling the black-box prediction API \cite{Tramr2016StealingML,CorreiaSilva2018CopycatCS,Orekondy2019KnockoffNS,Chandrasekaran2020ExploringCB} or directly reverse-engineers the model parameters by exploiting the property of the rectified linear units \cite{Oh2018TowardsRB,Jagielski2020HighAA,Carlini2020CryptanalyticEO}. With finer attack granularity, another branch of attacks aims at breaking the privacy of single training samples. For instance, \textit{membership inference attack} instead turns to infer whether a query sample belongs to the training set via white-box access \cite{Shokri2016MembershipIA}, by knowing the predicted confidence values \cite{Salem2019MLLeaksMA}, or by label-only exposures \cite{Li2021MembershipLI,ChoquetteChoo2021LabelOnlyMI} from a trained model. Differently, recent years further witness works that infer sample-level sensitive information or even reconstruct raw training samples from intermediate computation results (e.g., features \cite{Pan2020PrivacyRO} and gradients \cite{Melis2019ExploitingUF,Zhu2019DeepLF,Geiping2020InvertingG}) and model outputs \cite{Carlini2019TheSS,Salem2020UpdatesLeakDS,Radhakrishnan2020OverparameterizedNN,Carlini2021ExtractingTD}. In contrast, instead of exploiting a well-trained DNN for probing its training data, we for the first time explore and reveal the risk of a private training dataset with even no exposed interface, which we hope can substantially advance more comprehensive access control mechanisms.

\noindent\textbf{Data Hiding in Neural Networks.}
 Data hiding, or steganography, is a long-standing research area in our community \cite{Provos2003HideAS,Cheddad2010DigitalIS}. With the recent development of open-source model supply chains, several research works also exploit DNN as a new medium for hiding binary information. At the early stage, Uchida et al. \cite{Uchida2017EmbeddingWI} and Song et al. \cite{Song2017MachineLM} concurrently explore the idea of hiding data in DNN yet with different research focuses. To protect the intellectual property of DNN, Uchida et al. propose to embed a secret random binary message into the model parameters via conventional steganography techniques (e.g., least-significant bits or sign encoding). By verifying whether a model contains the binary message, the ownership is established. Later, this seminal work catalyzes the orthogonal study of DNN watermarking \cite{Adi2018TurningYW,Chen2019DeepMarksAS,Wang2021RIGACA}. Instead of model protection, Song et al. aim at hiding sensitive information about the private data into the model parameters during its training. Specifically, they directly convert a subset of sensitive data inputs into a binary form and encode them into the model parameters again with almost the same set of conventional steganography techniques in \cite{Uchida2017EmbeddingWI}.

\section{Preliminary}
\subsection{Background on Machine Learning (ML)}
\noindent\textbf{Notions in ML.}
In our work, we call a private
database is a \textit{ML dataset} if it is prepared for training ML models. In this paper, we mainly focus on supervised learning task, which is defined in a space $\mathcal{X}\times{\mathcal{Y}}$. Here, $\mathcal{X}$ is the input space, composed of raw data including but not limited to images, texts, or audios, and $Y$ is the label space. A learning model $f(\cdot;\theta) (:=f_\theta):\mathcal{X}\to\mathcal{Y}$ aims to build the relation from an element in $\mathcal{X}$ to the label in $\mathcal{Y}$, which consists of all the possible values in the annotation. For example, in a $K$-class image classification task, $\mathcal{X}$ consists of a set of images to be classified, while $\mathcal{Y} = \{1, \dots, K\}$, i.e., the possible classes.  Finally, the training process of a learning model $f_\theta$ involves the optimization of a \textit{loss function}, denoted as $\ell(\cdot; \theta)$, with respect to the parameters of the learning model.

\noindent\textbf{Basics of Deep Learning.}
\label{sec:prelim:dl}
The past decade has witnessed the evolution of deep learning (DL) techniques in most important ML tasks \cite{Goodfellow-et-al-2016}, which show substantial improvements over the classical ML techniques, and have already reach deep into a wide range of mission-critical applications in real world \cite{Heaton2016DeepLF,Esteva2017DermatologistlevelCO,Redmon2016YouOL}. The core of DL is \textit{deep neural networks} (DNNs), a family of learning models which are composed of a number of processing layers (e.g., linear/activation layers, convolutional/pooling layers, etc.) that transform the data representation through multiple abstraction levels \cite{Goodfellow2015DeepL}. Usually, a layer in DNN is composed of a set of \textit{learnable parameters}, which are iteratively updated to optimize the loss function via off-the-shelf gradient-based optimizers (e.g., Adam \cite{Kingma2015AdamAM}). 

As a final remark, we highlight proper parameter initialization is indispensable to attain a near-optimal DNN model throughout the training. Learnable parameters of different roles (i.e., weight, bias and scale) are usually initialized with distinct schemes and may also differ in the constraints on their values. For example, a weight scalar is usually initialized from Guassian, a bias scalar is initialized as $0$, while a scale scalar is $1$ with a nonnegative constraint. This technical detail influences our subsequent attack designs. 


\subsection{Data Security Models in AI Corporations}
\label{sec:data_security_study}
As essential backgrounds for discussing insider attacks, we provide below a field study on typical data security mechanisms in AI corporations. 

\noindent\textbf{Common Mechanisms}. Data Leakage Prevention Systems (DLPS) are one of the most mainstream security solutions in industry. DLPS can identify, monitor and protect confidential data and detect illegal data transmission based on predefined rules \cite{Alneyadi2016ASO}. According to Securosis, DLPS which are practically deployed inside corporations usually have rigorous restriction on data transfer to unauthorized endpoint devices (e.g., Wi-Fi, Bluetooth, USB) or to the outside network without permission \cite{securosis_report}, i.e., it is impossible for a naive insider attacker to directly transmit private datasets from the local network to the outside without being detected. With DLPS blocking potential data leakage to the outside, various access control policies are implemented by AI corporations to manage their owned private datasets. In the following, we introdce two typical modes named as the \textit{Willing-to-Share} mode and the \textit{Application-then-Authorization} mode, according to our industry partners.

\noindent\textbf{\textbullet\ The \textit{Willing-to-Share} Mode.} For start-up corporations which focus on one killer application (e.g., object detection \cite{Ren2015FasterRT}), they prefer to organize the datasets in a more shareable mode. According to our field study, a majority of the groups organize all the datasets (either from public or private sources) in their local distributed file system, which enables fast access to specific datasets according to one's requirement and facilitates swift development of novel DL techniques.

\noindent\textbf{\textbullet\ The \textit{Application-then-Authorization} Mode.}
More established AI corporations prefer to implement a more conservative way of managing ML datasets. According to Google's common security whitepaper \cite{google_security}, each employer should first apply for the authorization before accessing certain data resources, including the access to private datasets. However, considering the ever-evolving paradigms in deep learning, employees with ulterior motives may fabricate reasons such as the requirements of data augmentation \cite{Shorten2019ASO} or the purpose of multimodal learning \cite{Baltruaitis2019MultimodalML} to apply for relevant and irrelevant private datasets, which is common in social engineering \cite{Krombholz2015AdvancedSE}. Under the umberalla of DLPS, corporations may be less precautious about the unnecessary access to certain private datasets, as DLPS ought to forbid any attempts of transferring the private datasets away from the local network.     

\begin{tcolorbox}
\textbf{Summary}: An insider has multiple ways to access private datasets, which are strictly forbidden from being transferred out of the local network with no authorization. 
\end{tcolorbox}

\section{Security Settings}

\noindent\textbf{Attack Definition.} Following existing works on breaking training data privacy via well-trained DNN models (\S\ref{sec:related}), we define a target private dataset is \textit{stolen} in the sense of \textit{functionality leakage}, i.e., the attacker attains a trained model which has nearly the same functionality as if the model were trained on the target private dataset.

\noindent\textbf{Threat Model.} Our Matryoshka attack involves an inside attacker and an outside colluder. The colluder can also be played as the attacker him/herself. Formally, we have the following assumptions on the ability of the insider and the outsider: (i) \textit{Accessible Targets}: The insider has access to the target private dataset(s); (ii) \textit{Existent Carrier}: The insider is assigned with the task of training a DNN, which is scheduled to undergo an open sourcing process, on a non-target dataset; (iii) \textit{Receivable Carrier}: The outsider knows which model to download after the publishing; (iv) \textit{Secure Collusion}: The insider and the outsider can collude on several integer values and model architectures via a secure channel (e.g., a rendezvous).  

Below, we briefly discuss the rationale behind our threat model. First, \textit{the accessibility of the target datasets} is supported by our field study in \S\ref{sec:data_security_study}, where an insider can either freely access the target private dataset in the Willing-to-Share mode, or pretend to have a need on the target datasets to accomplish the assigned task at hand in the Application-then-Authorization mode. Second, \textit{the existence of a carrier} is realistic in the AI industry nowadays, where the trend of open sourcing pretrained DNNs becomes increasingly common for corporations with either educational or commercial purposes \cite{google_bert,openai_git}. Working as a DL engineer, the insider has a chance to be assigned with such a task. Besides, the third and the fourth assumptions can be easily fulfilled by off-line communication between the inside attacker and the outside colluder during non-working time. As we later show, in Matryoshka the secret keys which are required to decode the hidden information from the carrier model only consist of: a website link to the official downloading site, several integer values (i.e., random seeds) and the architectures of the secret models which can be chosen as well-known architectures that already standardized in popular DL frameworks to ease the collusion. This allows the collusion to be conducted via a rendezvous, which establishes a secure channel.

\noindent\textbf{Attack Taxonomy and Goals.}
To realize the covert transmission of secret models via a scheduled-to-publish DNN, our proposed Matryoshka attack is expected to meet the following attack goals: (i) \textit{Capacity:} As a prerequisite of invoking covert transmission, we require the secret models can be encoded into the carrier model without causing obvious degradation of its normal utility. Intuitively, a hiding scheme has higher capacity if it encodes more secret information with the same level of performance degradation. (ii) \textit{Decoding Efficiency:} This requirement measures how much additional knowledge is required to successfully decode the secret models from the carrier model. If it surpasses the memorization ability of the insider, the additional knowledge should be communicated via an additional round of covert transmission, increasing the risk of being detected. (iii) \textit{Effectiveness:} After the secret information is decoded from our carrier model, the information is expected to have almost no distortion (i.e., effective transmission). Specifically, in our attack, we require the decoded secret models are effective in stealing the target private dataset(s) in terms of leaking functionality or leaking sensitive data. (iv) \textit{Robustness:} Using the language of information theory, this requirement expects the covert transmission remains effective even if the transmission channel has noises. In our context, we require the colluder can still decode the secret models and thus the privacy of the target datasets when the carrier model undergoes common post-processing techniques (e.g., pruning and finetuning). (v) \textit{Covertness:} Finally, to reify the requirement of covertness, a third party should not be able to detect whether a published model contains secret models or not. Otherwise, the transmission would be canceled and the inside attacker will be detected according to the logging of dataset access.  

\begin{figure*}[t]
\begin{center}
\includegraphics[width=1.0\textwidth]{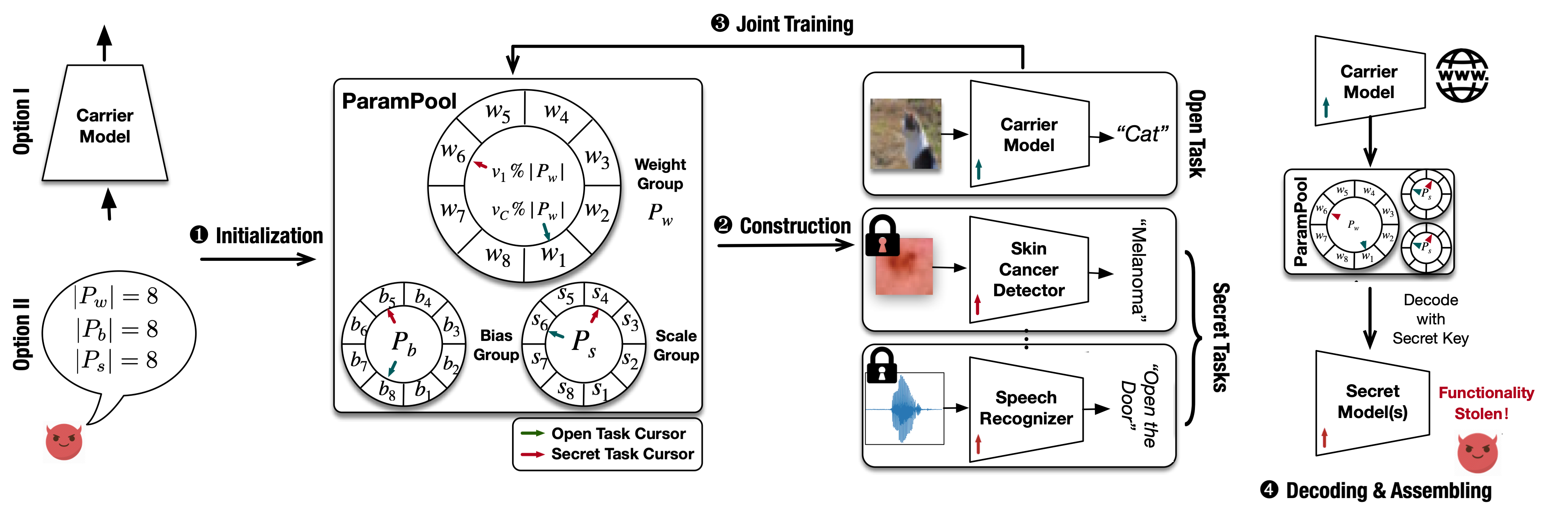}
\vspace{-0.2in}
\caption{Overview of the methodology of Matryoshka.}
\label{fig:atk_pipeline}
\end{center}
\vspace{-0.2in}
\end{figure*}

\noindent\textbf{Existing Data Hiding Techniques for Model Hiding.}
Data hiding is a long-standing and fruitful research area in security \cite{Provos2003HideAS}. Many well-known algorithms and tools are devised to encode binary messages in multimedia contents such as image \cite{Cheddad2010DigitalIS}, text \cite{Yang2019RNNStegaLS}, and audio \cite{Djebbar2012ComparativeSO}. Recently, there are also several research works which explore applying well-established data hiding techniques, including least-significant-bit (LSB \cite{Kurak1992ACN}), correlated value and sign encoding, to embed secret binary watermark \cite{Uchida2017EmbeddingWI,Wang2021RIGACA}, a subset of training data in binary form into a trained DNN \cite{Song2017MachineLM} or a virus \cite{Liu2020StegoNetTD}. 

However, existing data hiding techniques mostly view the carrier medium, whether common multimedia contents or parameters in DNN, as bit strings for encoding the secrets. Hence, the size of the carrier medium poses a strict capacity limit on the information that can be hidden. On the one hand, \textit{common multimedia contents can hardly afford to hide a DNN} which is usually composed of millions of floating-point parameters. For example, the storage of a ResNet-18 for ImageNet is over $45$MB, while, for image, the maximally tolerable embedding rate is only $0.4$ bits per pixel \cite{bossbase}. On the other hand, \textit{applying existing techniques to hide information in a carrier DNN cannot fully exploit its potential capacity.} Compared with conventional multimedia, a carrier DNN is much larger than traditional multimedia and has stronger resilience against the modification of multiple LSB positions \cite{Jacob2018QuantizationAT,Krishnamoorthi2018QuantizingDC}. Despite this, previous encoding schemes are only able to use about $20\%\sim{50\%}$ size of a DNN for information hiding. For example, Song et al. find most existing data hiding techniques can hardly hide over $500$ raw gray-scale images of $32\times{32}$ resolution ($5.85$MB in total) if the performance degradation of a carrier ResNet-18 as is required to be bound by $3\%$ \cite{Song2017MachineLM}. In our viewpoint, to directly encode raw data inputs into the carrier model confront the challenges from the distribution misalignment between the two heterogeneous data types (\S\ref{sec:pilot_study}). If existing techniques are applied for covert transmission of secret models, the flexibility and efficiency of our Matryoshka attack would be severely hurt: Either a larger secret model or multiple secret models for different attack purposes are unable to be hidden in the carrier model. 
\begin{tcolorbox}
\textbf{Summary}: (i) Multimedia carrier usually have insufficient capacity for model hiding. (ii) Existing techniques have not fully exploited the encoding capacity of a carrier DNN.  
\end{tcolorbox}

\section{The Matryoshka Attack}

\label{sec:atk}
\subsection{Attack Overview}
 Below, we first present our design and insights of a \textit{ParamPool}, which is at the core of the encoding (decoding) of the secret models into (from) the carrier model.

\noindent\textbf{Design of ParamPool.} In general, a ParamPool $P$ maintains an array of $|P|$ learnable scalar parameters. Corresponding to the different types (e.g., weight, bias, scale) of learnable parameters in DNN, the parameters in a weight pool are also grouped in disjoint groups, i.e., $P = P_{w} \cup P_b \cup P_{s}$. Each group implements different random schemes when initialized. Given an integer secret $v$, a ParamPool $P$ implements the following primitives to interact with a DNN $f(\cdot; \theta)$:
\begin{itemize}
    \item \textit{(i)} \textbf{Fill}($P$, $f$, $v$) $\to {\mathcal{P}(\cdot, v)}$: This primitive replaces each original parameter in $f$ with a parameter of the same type which is selected from the ParamPool $P$ with \textit{replacement} according to the secret key $v$. We use $\mathcal{P}(\theta, v)$ to denote the parameters of a DNN $f$ filled by $P$ under the secret integer $v$. In this sense, $\mathcal{P}(\cdot, v)$ is viewed as a hash map from the original parameters to the ParamPool parameters. 
    \item \textit{(ii)} \textbf{Propagate}($P$, $f(\cdot; \theta)$, $v$) $\to \emptyset$: After an optimization step on $f$, this primitive collects the weight update on each parameter in a DNN and propagates them to the corresponding positions in the ParamPool with  $\mathcal{P}(\cdot; v)$. 
    \item \textit{(iii)} \textbf{Update}($P$)$\to \emptyset$: After the weight update is accumulated in the \textit{update buffer} of the corresponding parameter in $P$, this primitive updates each parameter in $P$ by aggregating its update buffer and then reset it.
    \item \textit{(iv)}  \textbf{Decode}($f$, $v$)$\to P$: This primitive decodes the the ParamPool from a DNN $f$ according to the secret key $v$. 
\end{itemize}
Intuitively, with the above primitives, a ParamPool can be used as a proxy to the full life cycle of a DNN.

\noindent\textbf{Attack Pipeline.}
\label{sec:attack_pipeline}
Fig.\ref{fig:atk_pipeline} provides an overview of our attack pipeline, which is mainly divided into four stages. In the following, we denote the carrier DNN as $C$.

\noindent$\bullet$\textbf{ Stage 1: Initialization of ParamPool.} First, a ParamPool $P$ is initialized from the carrier DNN by colllecting and grouping all the learnable parameters \textit{(Option I)} or from scratch with an attacker-specified size for each group \textit{(Option II)}. Later we show Option I is more advantageous in evading detection (\S\ref{sec:evasion}) while 
    Option II helps the carrier model to be more robust even transmitted in a noisy channel, i.e., model post-processing (\S\ref{sec:exp:resilience}).
      
\noindent$\bullet$\textbf{ Stage 2: Construction of the Secret Tasks} (\S\ref{sec:atk:morph}). In the next stage, we specify the secret learning tasks according to the attack purposes and the nature of the target datasets $D_1, \hdots, D_N$ to steal ($N\ge{1}$). For stealing the functionality of a supervised learning dataset, we choose an off-the-shelf DNN architecture to fit the private dataset with a proper loss $\ell_k$. Differently, for stealing a sensitive subset of private data inputs, we choose a DNN which is trained to map a sequence of noise vectors to the attacker-interested data inputs in a one-to-one way. The noise vectors are randomly sampled from a secret distribution with a fixed integer seed, which is exclusively known to the insider and the colluder. Finally, the attacker chooses an integer secret $v_k$ and invokes the \textbf{Fill} primitive to replace the parameters in each secret model with the ones in $P$ by invoking \textbf{Fill}($P$, $f_k$, $v_k$).

\noindent$\bullet$\textbf{ Stage 3: Joint Training for Model Hiding} (\S\ref{sec:atk:joint}) Combining with the learning task $(D_C, \ell_C)$ of the carrier model $C$, the attacker jointly trains the carrier model and the secret models by optimizing the parameters in the ParamPool $P$. Concisely, in each iteration, we synchronously calculate the parameter updates in each model and then invoke \textbf{Propagate}($P$, $f_k$) to accumulate the updates in each model to the corresponding update buffers. Subsequently, we invoke the \textbf{Update} primitive and resume the joint training to the next iteration. When the training finishes, the attacker removes all the traces in the code base and replaces the corresponding parameters in the carrier model with the values from the final ParamPool. We denote the final carrier model as $C^{*}$. 

\noindent$\bullet$\textbf{ Stage 4: Decoding Secret Models from the Carrier Model} (\S\ref{sec:atk:decode}). After the carrier model $C^{*}$ is published, the outsider colluder downloads $C^{*}$. With the knowledge of the architectures of the secret models and the secret key $v_k$ communicated via a secure channel, the outsider first invokes \textbf{Decode}($C^{*}$, $v_k$) to decode the ParamPool $P$ from the carrier model. Then, the colluder assembles the secret models from the decoded ParamPool. In the following, we present the detailed technical designs for each stage above.

\subsection{Construction of the Secret Tasks}
\label{sec:atk:morph}
\noindent\textbf{Functionality-Oriented Learning Tasks.} For stealing the functionality of a private dataset, the secret learning task is constructed as if the attacker is training a usable model $f_k(\cdot;\theta_k):\mathcal{X}_k\to\mathcal{Y}_k$ on the private dataset $D_k := \{(x_i, y_i)\}_{i=1}^{M_k}$ belonging to $\mathcal{X}_k\times\mathcal{Y}_{k}$. To minimize the required information for collusion, we propose to use standardized neural architectures in deep learning frameworks. With the loss function $\ell_{k}$, we formally construct the secret learning task for functionality stealing as 
$   \min_{\theta_k} L_k(\theta_k) := {1}/{|D_k|} \sum_{(x_i,y_i) \in D_k} \ell_k(f_k(x_i; \theta_k), y_i)$.

\noindent\textbf{\textbf{Fill} Secret/Carrier Models with ParamPool.} With the carrier model $C$ and an initialized ParamPool $P$, we first specify a random integer $v_k$ for each secret model $f_k$ from all the possible indices of the ParamPool $P$, i.e., $\{1, 2, \hdots, |P|\}$. This prevents the secret models from being decoded by any party except for the colluder. Then we invoke the primitive \textbf{Fill}($P$, $f_k$, $v_k$) to replace the original parameters in $f_k$ by parameters in $P$. 
Intuitively, the \textbf{Fill} primitive loops over all the scalar parameters in the target model $f_k(\cdot; \theta)$ and assign it with the value of a parameter selected from the ParamPool. As Fig.\ref{fig:atk_pipeline} shows, the parameter selection cursor on each parameter group (e.g., the weight group $P_w$) cycles from a starting index derived from the integer secret (e.g., $v_k\mod |P_w|$). For the carrier model, we choose $v_C=0$ if the ParamPool is initialized directly from the carrier model (i.e., Option II in \S\ref{sec:attack_pipeline}), or otherwise, we sample an integer secret $v_C$ as well for the carrier model.  The obtained model is denoted as $\tilde{f}_k$ with the substituted parameters as $\mathcal{P}(\theta_k, v_k)$.


\subsection{Joint Training for Model Hiding}
\label{sec:atk:joint}
After the secret and the carrier models are filled, the ParamPool $P$ is ready to become a proxy to the training process of each secret/open learning task. Without loss of generality, the carrier model $C$ is supposed to be trained on a supervised learning task $D_{C} := \{(x_i, y_i)\}_{i=1}^{M_C}$ with a loss function $\ell_C$. Formally, the model hiding process aims to solve the following joint learning objective: 
\begin{align}
    \min_{P} \frac{1}{|D_C|} \sum_{(x_i,y_i) \in D_C} \ell_k(C(x_i; \mathcal{P}(\theta_C, v_C)), y_i) \nonumber \\ + \frac{1}{N}\sum_{k=1}^{N}L_k(\mathcal{P}(\theta_k, v_k)).
\end{align}
Intuitively, the above objective requires $P$ to reach a good consensus on $N$ secret tasks and the open task. For example, when $N = 1$, it means that the sets of local optimum for $f_C$ and $f_1$ should intersect with one another to guarantee a near-optimal ParamPool is attainable. For the first time, we empirically propose a joint training algorithm in Algorithm \ref{alg:train} which solves the above learning objective to construct a near-optimal ParamPool. Each secret model assembled from the optimized ParamPool exhibits similar utility compared with an identical model which is independently trained (\S\ref{sec:exp:multiple}). However, to the best of our knowledge, existing deep learning theories can hardly help with the explanation of this phenomenon. To provide a tentative explanation, we provide the following pilot study on the optimal parameter distributions across models.

\begin{figure}[t]
\begin{center}
\includegraphics[width=0.5\textwidth]{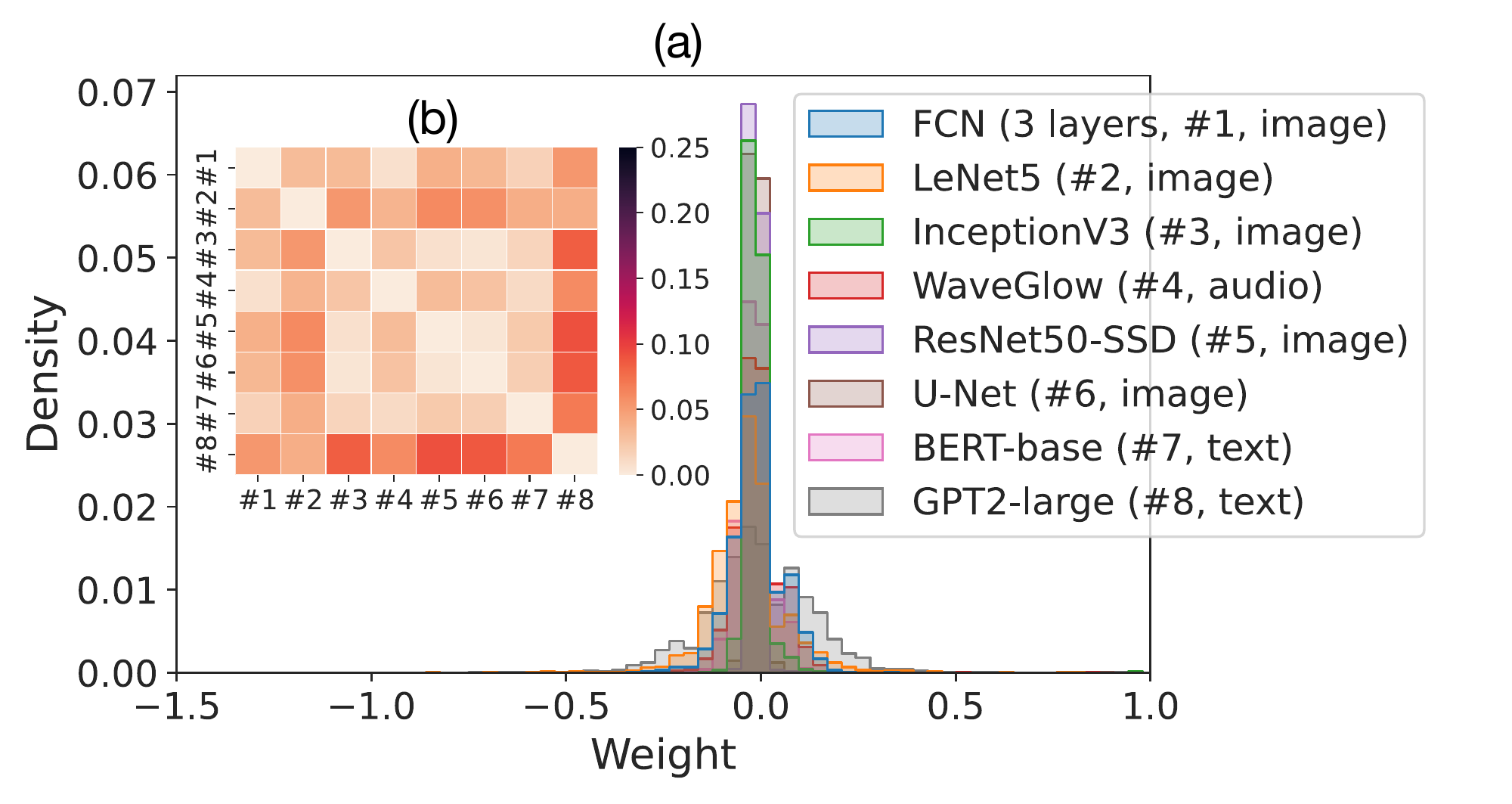}
\vspace{-0.2in}
\caption{(a) The weight histogram of $8$ pretrained models of high diversity. (b) The empirical values for the pairwise OTD among the $8$ models.}
\label{fig:pilot_study}
\end{center}
\vspace{-0.3in}
\end{figure}

\noindent\textbf{A Pilot Study on Parameter Distribution.}
\label{sec:pilot_study} Specifically, we download $8$ pretrained models (i.e., $\{f_k(\cdot; \theta_k \}_{k=1}^{8}$) from Pytorch Hub \cite{apple_hub}, with their names listed in the legend of Fig.\ref{fig:pilot_study}. The selected models cover typical applications (e.g., classification, generation, object detection) and span image, audio and text data.  In the following, we focus on the parameters of the \textit{weight} type. Without loss of generality, similar arguments are applicable to the bias and the scale. First, we calculate the normalized histogram of the parameters of the \textit{weight} type in each model on a range of $[-1, 1]$ with $n=100$ bins. In our experiments, the range is validated to cover all the weight parameters. We plot the weight histograms in Fig.\ref{fig:pilot_study}(a) and use $H_w(\theta_k) := \{(w_1, p_{1}(\theta_k)), \hdots, (w_n, p_{n}(\theta_k))\}$ to denote the histogram of the weights in $\theta_k$. To analyze the feasibility of our ParamPool-based weight sharing scheme, we measure the following optimal transportation distance (OTD) between the weight histograms $H_w(\theta_j)$ and $H_w(\theta_k)$ \cite{Rubner2004TheEM}: 
\begin{align}
    & \text{OTD}(H_w(\theta_j), H_w(\theta_k)) = \min  \sum_{l=1}^{n}\sum_{m=1}^{n}d_{lm}p_{lm} \\
    & \text{s.t., } \sum_{l=1}^{n}p_{lm} \le p_{m}(\theta_k), \sum_{m=1}^{n}p_{lm} \le p_{l}(\theta_j), \sum_{m=1}^{n}\sum_{l=1}^{n}p_{lm} = 1,
\end{align}
where $d_{ij} = |w_{i} - w_{j}|$, i.e., the distance between the $i$-th and the $j$-th bin centers. We call the $\{p_{jk}^{*}\}$ to solve the linear programming above as \textit{the optimal transportation (OT) scheme}. In general, the OTD between two distributions measures the minimal distance when the density in one distribution is transported to another \cite{villani2008optimal}. In our context, viewing the ParamPool as the intermediate of mapping a model $f_j$ to a different model $f_k$, we present the following analytical result.
\begin{theorem}[Existence of Near-Optimal ParamPool] For a pair of learning tasks $(D_j, f_j, \ell_j)$ and $(D_j, f_j, \ell_j)$ such that $\theta_j^{*}$ and ${\theta_k}^{*}$ are the corresponding minimizers to the training loss, if $\text{OTD}(H_{\cdot}(\theta_i), H_{\cdot}(\theta_j)) < \epsilon$ (where $\cdot$ iterates in $\{w, b, s\}$, i.e., weight, bias, scale), then there always exist a ParamPool $\mathcal{P}$ of size $|P|$ and two integer secrets $v_j, v_k$ such that
\begin{align}
    |\mathcal{P}(\theta_j^{*}, v_j)[i] - \theta_j^{*}[i]| < \epsilon/2 + O({|P|}^{-1}) \nonumber \\ |\mathcal{P}(\theta_j^{*}, v_k)[i] - \theta_k^{*}[i]| < \epsilon/2 + O({|P|}^{-1})
\end{align}
where $\theta_j^{*}[i]$ denotes the $i$-th scalar parameter in $\theta_j^{*}$.
\label{thm:existence_optimal}

\end{theorem}

In plain words, the above theorem argues that the existence of a near-optimal ParamPool which minimizes the loss functions on both individual task is strongly related with the OTD between the parameter distributions. Intuitively, when an $\text{OTD}(H(\theta_j), H(\theta_k))$ is smaller than $\epsilon$, it means that a scalar weight parameter $w$ in $f_j$ can always find a substitute parameter $w'$ such that $|w - w'| < \epsilon$ due to the existence of the OT scheme. Therefore, with the parameters of $f_j, f_k$, one can always construct a ParamPool and the mapping relations from the models to the ParamPool such that the shared weight differs from the optimized weight by $\epsilon$. The following provides the proof for this theorem.

\begin{proof}[Proof for Theorem \ref{thm:existence_optimal}]
First, we provide a constructive proof to the case when the ParamPool has an unlimited size, which is formally stated as the following lemma. 
\begin{lma}
If $\text{OTD}(H_{\cdot}(\theta_i), H_{\cdot}(\theta_j)) < \epsilon$ (where $\cdot$ iterates in $\{w,b,s\}$), then there always exist a ParamPool $P'$ and two integer secrets $v_j$ and $v_k$ such that, for each $i$, $
    |\mathcal{P}'(\theta_j^{*}, v_j)[i] - \theta_j^{*}[i]| < \epsilon/2, |\mathcal{P}'(\theta_k^{*}, v_k)[i] - \theta_k^{*}[i]| < \epsilon/2. $
\label{lma:parampool_constr}
\end{lma}
\begin{proof}
By the definition of $\text{OTD}(H_{\cdot}(\theta_i), H_{\cdot}(\theta_j)) < \epsilon$, there exists an optimal transportation scheme $p_{jk}^{*}$ such that for each $\theta_{j}^{*}[i]$, the corresponding parameter $\theta_{k}^{*}[p_{jk}^{*}(i)]$ lies in the $\epsilon$-ball of $\theta_{j}^{*}[i]$. Therefore, for the index $i$ of the parameter $\theta_{j}^{*}$ and the index $p_{jk}^{*}(i)$ of the parameter $\theta_{k}^{*}$, the $l$-th ParamPool parameter $P'[l]$ can always be constructed to satisfy $|\theta_{j}^{*}[i] - P'[l]| < \epsilon/2$ and $|\theta_{k}^{*}[p_{jk}^{*}(i)] - P'[l]| < \epsilon/2$ (e.g., by choosing $P'[l] = (\theta_{j}^{*}[i] + \theta_{k}^{*}[p_{jk}^{*}(i)])/2$). Therefore, a mapping rule from $\theta^{*}_i$ to $P'$ is $i\to{l}$, while, for $\theta^{*}_j$, it is $p_{jk}^{*}(i)\to{l}$. Without loss of generality, we assume $\theta_i^{*}$ is no smaller than $\theta_{j}^{*}$. By iterating over $i$ as above, we construct a ParamPool $P'$ of the same size as $\theta_{i}^{*}$, and the mapping relations $\mathcal{P}'(\theta_{i}^{*}, v_i) = \{i\to{l}\}$, $\mathcal{P}'(\theta_{j}^{*}, v_i) = \{p_{jk}^{*}(i)\to{l}\}$. One can easily show there exist integers $v_i, v_j$ conforming to the above mapping relation.
\end{proof}
If the size of $\theta_{j}^{*}$ is no smaller than $|P|$, our theorem is already valid according to the lemma above. Otherwise, we can extend the above lemma to the case by clustering the parameters in the unlimited ParamPool we construct above to satisfy the size constraint. Technically, considering the ParamPool $P'$ in Lemma \ref{lma:parampool_constr}, we next prove the existence of a smaller ParamPool of size $|P|$ which well approximates $P'$ in the following sense. 
\begin{lma}
There exists a ParamPool $P$ and a mapping relation $\pi$ from the index of $P'$ to the index of $P$ such that $|P'[i] - P[\pi(j)]| < O(|P|^{-1})$.
\end{lma}
\begin{proof}
We prove the lemma again via construction. By dividing the parameter range (w.l.o.g., we denote the range as $[-M/2, M/2]$) in $|P|$ bins of equal width (i.e., $M/|P|$), we put the parameters in $P'$ into each bin. Then we construct the smaller ParamPool by collecting all the $|P|$ bin centers, and the mapping relation as the belonging relation to the bins. The the lemma is proved. 
\end{proof}

Finally, we leverage the triangle inequality to such that $|\theta_{j}^{*}[i] - P[l]| < |\theta_{j}^{*}[i] - P'[l]| +  |P'[l] - P'[\pi(l)]| < \epsilon/2 + O(|P|^{-1})$.  Finally, by contracting on the original mapping relations in $\mathcal{P}'(\theta_{j}^{*}, v_i)$ with the clustering relation $\pi$, we construct the mapping relation from $\theta_{j}^{*}$ to the ParamPool $P$. Similar arguments hold for $\theta_{k}^{*}$. 
\end{proof}


Finally, we empirically calculate the pairwise OTD among the $8$ models based on the histogram, and report the results in the heatmap form in Fig.\ref{fig:pilot_study}(b). The results show, the upper bound $\epsilon$ is $0.038$ on average with a standard deviation of $0.008$ (i.e. the $\epsilon$ is smaller than $2\%$ of the weight range of length $2$), which would hardly affect the model performance \cite{Weng2020TowardsCM}. To summarize, our pilot study relates the existence of a near-optimal ParamPool to the scale of the OTD between the distribution of the optimized parameters. For future exploration, we leave open a more comprehensive theoretical treatment on how such a near-optimal ParamPool is possible to be attained with our proposed algorithm.

\noindent\textbf{Searching for the Optimized ParamPool.}
To solve the joint learning objective in Section \ref{sec:atk:joint}, our proposed attack executes the following training iteration recurrently. Denote the ParamPool at the $t$-th iteration as $P_t$. Concisely, at the $t$-th iteration, we iterate over all the $N$ secret tasks and the normal task to conduct the following key procedures:
\begin{itemize}
\item \textit{(i)} For the $k$-th secret tasks, we first invoke \textbf{Fill}($P_t$, $f_k$, $v_k$) to instantiate $f_k$ with the current values of the ParamPool. 
\item \textit{(ii)}  Then, we forward a training batch via the model $f_k(\cdot; P_t(\theta_k, v_k))$, back-propagate the loss $L_k$ approximated on the training batch, and conduct one optimization step on the parameters of $f_k$ with an optimizer (e.g., Adam \cite{Kingma2015AdamAM}). 
\item \textit{(iii)} Finally, we collect the weight update on each parameter and follow the mapping relation in $\mathcal{P}(\cdot, v_k)$ to propagate the update to the corresponding ParamPool parameter. The above procedures are also conducted on the carrier model.
\end{itemize}
The above procedures describe the \textbf{Propagate} primitive on each secret/open task (L4-15 in Algorithm \ref{alg:train}).

 
 \begin{algorithm}
 \caption{The $t$-th iteration during the joint training process on the ParamPool}
 \label{alg:train}
 { 
 \begin{algorithmic}[1]

 \renewcommand{\algorithmicrequire}{\textbf{Input:}}
 \renewcommand{\algorithmicensure}{\textbf{Output:}}
 \REQUIRE $P_t$ (the current ParamPool), $\{(f_k, \tilde{D}_k, \ell_k, \text{Opt}_k\ )\}_{k=0}^{N}$ (the secret and open tasks).
 {\small{\hskip1em$\vartriangleright$ For simplicity, the index $0$ denotes the open task on the carrier model.}} 
\ENSURE $P_{t+1}$ (the updated ParamPool)
 \FOR{each parameter $w_P$ in $P_t$}
    \STATE $w_P$.buffer $\gets \{\}$ \\ 
 \ENDFOR
  \PARALLEL{$k=0, 1, \hdots, N$}
\begin{ALC@g}
  \STATE $\mathcal{P}(\cdot, v_k) \gets$
  \textbf{Fill}($P_t$, $f_k$, $v_k$) \\
  \STATE Sample a training batch $B$ from $\tilde{D}_k$. \\
  \STATE $\tilde{L}_k \gets \frac{1}{|B|}\sum_{(x,y)\in{B}}\ell_k(f_k(x; \mathcal{P}(\theta_k, v_k)), y)$.
   {\small{\hskip0em$\vartriangleright$ Approximate the loss $L_k$ on a randomly sampled mini-batch.}} \\
   \STATE $\tilde{L}_k$.Backward() \\
   \STATE $\Delta{\theta_k} \gets \text{Opt}_k$.\text{Step}() \\ 
  \FOR{each scalar parameter $w$ in $\theta_k$}
    \IF{$w$ belongs to a \{weight, scale, bias\} parameter}
    \STATE $\mathcal{P}(w, v_k)$.buffer.Append($\Delta{w}$)
    {\small{\hskip0em$\vartriangleright$ Propagate the update to the corresponding ParamPool parameter.}} \\
    \ENDIF
  \ENDFOR
\end{ALC@g}
  \ENDPARALLEL
 \FOR{each parameter $w_p$ in $P_t$}
    \STATE $w_p \gets w_p + $ Average($w_p$.buffer) \\ 
 \ENDFOR
 \RETURN $P_{t+1}$.
 \end{algorithmic}}
 \end{algorithm}

The final step in one training iteration is to invoke the \textbf{Update} primitive on the ParamPool (L16-18 in Algorithm \ref{alg:train}). Technically, for each parameter in $P_t$, we maintain a buffer to store the weight updates from each task. The \textit{update buffer} is aggregated to obtain the global update value on the corresponding scalar parameter in $P_t$. In our experiments, we find aggregation by average is already sufficient to achieve effective attacks. After the parameter is updated, we clear the update buffers and resume the next training iteration. 

 As a final remark, the training process on the ParamPool terminates when (i) the validation performance of the carrier model meets the task requirement, and (ii) at least one of the secret tasks converge. This condition ensures the carrier model to smoothly transition to the next stage of a publishing process with certain secret models encoded. We denote the final ParamPool as $P^{*}$. To wind up the model hiding phase, the attacker clears up any traces of malicious training code and irrelevant intermediate outcomes, memorizes the secret keys (i.e., the random seeds $\{s_k\}$ for generating noises, the starting indices $\{v_k\}$ and the architecture name for each task, the size of each group in the ParamPool) via a secure medium, and instantiates the carrier model by \textbf{Fill}($P^{*}$, $f_C$, $v_C$).

\subsection{Decoding Secret Models from the Carrier Model}
\label{sec:atk:decode}
\noindent\textbf{Recovering the ParamPool.} After the carrier model is published online with open access, the attacker immediately notifies the outside colluder to download the carrier model and decode the secret models from the carrier model. Specifically, after colluding on the secret keys with the attacker via a secure channel (e.g., an in-person rendezvous), the outsider first decodes the ParamPool based on the colluded knowledge of the ParamPool sizes and the starting index $v_C$, i.e., by the primitive $\text{Decode}(C, v_C)$. Specifically, the decoding procedure differs according to the ParamPool initialization:
\begin{itemize}
\item \textit{(i)} If initialized from the carrier model directly, the ParamPool can be easily recovered by the colluder via collecting each parameter in the carrier model into different groups. 
\item \textit{(ii)} Otherwise, if the ParamPool is created from scratch, the colluder first collects the parameters of different groups from the carrier model. Then, the attacker slices, e.g., the weight parameters into segments of length $|P_w|$. The last segment may need additional zero padding to hold the same length.  Finally, the attacker conducts \textit{a fusion operation} on the $N$ decoded segments, right-shift the fusion result by $v_k \mod |P_w|$, and permute it with a permutation inverse to the one in \textbf{Fill} to recover the final $P_w$ in the ParamPool. Similar operations are conducted on the bias and the scale groups. We introduce the fusion mechanism on the $N$ decoded segments to implement resilience against post-processing on the carrier model (\S\ref{sec:exp:resilience}). For example, when the colluder finds the carrier model is pruned, the fusion mechanism selects the non-zero value from each ParamPool copy to restore the pruned values.

\end{itemize}
More details on the decoding procedure are provided in Algorithm \ref{alg:decode}. To determine which decoding algorithm to use, the outsider simply compares the known $|P|$ with the number of learnable parameters in the carrier model.

 \begin{algorithm}
 \caption{The \textbf{Decode}($C$, $v_c$) Primitive }
 \label{alg:decode}
  {\fontsize{10}{10}\selectfont
 \begin{algorithmic}[1]
 \renewcommand{\algorithmicrequire}{\textbf{Input:}}
 \renewcommand{\algorithmicensure}{\textbf{Output:}}
 \REQUIRE $C$ (the carrier model), $v_c$ (the integer secret of the carrier model), $N_w, N_b, N_s$ (the length of $P_w, P_b, P_s$).
 \ENSURE $P = P_w \cup P_b \cup P_s$ (the decoded ParamPool).
\STATE{$P_w\gets list(),P_b\gets list(), P_s\gets list()$}

 \FOR {each scalar parameter $w$ in $C$}
 \IF{$w$ belongs to a \{weight, scale, bias\} parameter}
 \SWITCH{$\text{type}(w)$}
  \CASE{weight}
   \STATE {$P_w$.Append($w$.data) }
  \ENDCASE
    \CASE{bias}
   \STATE  {$P_b$.Append($w$.data) } 
  \ENDCASE
    \CASE{scale}
   \STATE {$P_s$.Append($w$.data) }
  \ENDCASE
 \ENDSWITCH
  \ELSE
  \STATE{CONTINUE}
  \ENDIF
  \ENDFOR

\STATE {$v_w, v_b, v_s \gets v_c \mod N_w, v_c \mod N_b, v_c \mod N_s$}
\STATE Generate a random permutation $\pi_w$ ($\pi_b, \pi_s$) of integers from 0 to $N_w$  ($N_b, N_s$) with seeds $v_w (v_b, v_s)$.

  \STATE {$P_w, P_b, P_s \gets \text{Fusion}(P_w,  P_b,  P_s)$} 
  {\small{\hskip3em$\vartriangleright$ Recover the parameter by selecting the non-zero value from each ParamPool copy, if needed.}}
  \STATE Right-shift each parameter group by $v_w, v_b, v_s$.
 \STATE {$P_w, P_b, P_s \gets \pi_w ^{-1} \circ P_w, \pi_b ^{-1} \circ P_b, \pi_s ^{-1} \circ P_s$} 
 \RETURN  $P = P_w \cup P_b \cup P_s$
 \end{algorithmic}}
 \end{algorithm}

\noindent\textbf{Assembling the Secret Models.} Finally, the colluder recovers the secret models $f_1, f_2, \hdots, f_N$ by invoking the \textbf{Fill} primitive with the decoded ParamPool in the previous part. In this way, the attack objective is attained. 


\section{Evaluation Results}
\subsection{Overview of Evaluation}



\begin{table*}[htbp]
  \centering
  \caption{Effectiveness of our Matryoshka attack in transmitting single secret models via a ResNet-18 on CIFAR-10.}
  \scalebox{0.9}{
    \begin{tabular}{clccrrrrrc}
    \toprule
    \multirow{2}[4]{*}{\textbf{Domain}} & \multicolumn{1}{c}{\multirow{2}[4]{*}{\textbf{Task}}} & \multicolumn{2}{c}{\textbf{Secret Tasks}} & \multicolumn{3}{c}{\textbf{Secret Model}} & \multicolumn{3}{c}{\textbf{Carrier Model}} \\
 \cmidrule(lr){3-4} \cmidrule(lr){5-7} \cmidrule(lr){8-10}        &       & \multicolumn{1}{c}{\textbf{Dataset/Model/Metric}} & \multicolumn{1}{c}{\textbf{\# Params}} & \multicolumn{1}{l}{$\Delta$\textit{Perf}-\textbf{I}} & \multicolumn{1}{l}{$\Delta$\textit{Perf}-\textbf{II}} & \multicolumn{1}{l}{\textbf{Normal}} & \multicolumn{1}{l}{$\Delta$\textit{Perf}-\textbf{I}} & \multicolumn{1}{l}{$\Delta$\textit{Perf}-\textbf{II}} & \textbf{Normal} \\
    \midrule
    \multirow{2}[1]{*}{Image} & Classification & GTSRB\cite{Houben-IJCNN-2013}/VGG-16\cite{simonyan2014very}/ACC & 14.74M & -0.51\%  & 0.25\%  & 98.45\% & 0.03\%  & -0.61\%  & \multirow{7}[9]{*}{95.50\%} \\
          & Object Detection & VOC'07\cite{Everingham15}/MobileNetV1-SSD\cite{howard2017mobilenets}/MAP & 9.47M &
         -1.42\% & -16.63\% & 	51.57\% & 	0.12\% & 	-0.61\% & \\
\cmidrule{1-9}    Text  & Classification & IMDB\cite{maas-EtAl:2011:ACL-HLT2011}/TextCNN\cite{zhang2015sensitivity}/ACC & 6.69M & -5.66\%  & -9.08\%  & 90.85\% & -0.16\%  & -0.41\%  &  \\
\cmidrule{1-9}    Audio & Classification & SpeechCommand\cite{speechcommandsv2}/M5\cite{7952190}/ACC & 26.92K & -1.39\%  & -1.11\%  & 96.86\% & -0.32\%  & -1.14\%  &  \\
\cmidrule{1-9}    Time Series & Classification & UWave\cite{liu2009uwave}/RNN/MSE & 1.67K & -1.14\%  & -4.54\%  & 86.36\% & -0.67\%  & -0.40\%  &  \\
\cmidrule{1-9}    \multirow{2}[1]{*}{Healthcare} & Classification & DermaMNIST\cite{medmnistv2}/LeNet-5\cite{lecun1998gradient}/ACC & 61.98K & -4.78\%  & -5.74\%  & 74.21\% & -0.34\%  & -0.47\%  &  \\
          & Regression & Warfarin\cite{Truda2020warfit}/MLP/MSE & 0.5K  & -0.51  & -0.86  & 8.23  & 0.00\%     & -0.35\% &  \\
\bottomrule
    \end{tabular}}%
  \label{tab:carrier_suite}%
\end{table*}%


\noindent\textbf{Datasets \& Architectures.}
We evaluate the performance of Matroyoshka on a diversity of datasets and tasks from multiple application domains including image, audio, text, time series and healthcare. The general information is presented in \textit{Secret Tasks} columns of Table \ref{tab:carrier_multi}. 
Due to ethical concerns, we conduct all our experiments on public datasets. To faciliate future research, we open source our experimental code in \url{https://anonymous.4open.science/r/hiding_models-EAF8}.



\noindent\textbf{Evaluation Metrics.}
We measure the effectiveness of functionality-oriented stealing with \textit{Performance Difference} ($\Delta$\textit{Perf}). For the carrier model, $\Delta$\textit{Perf} measures the difference between the carrier model encoded with secret models and an independently trained carrier model. For a secret model, $\Delta$\textit{Perf} measures the performance difference between the decoded secret model and a model of the same architecture while independently trained on the private dataset. A lower $\Delta$\textit{Perf} means model hiding causes more performance overhead to the carrier model or the secret model, which implies the carrier model has lower capacity. Moreover, a lower carrier model $\Delta$\textit{Perf} means model hiding is less covert, while a lower secret model $\Delta$\textit{Perf} indicates the attack is less effective. Specifically, the $7$ secret tasks involve the following performance metrics:
\begin{itemize}
    \item \textbf{ACC}: Accuracy (ACC) is the standard evaluation metric for classification tasks. ACC measures the proportion of correctly predicted samples. It formally writes $\operatorname{ACC} = \frac{\sum_{i=1}^{N} \mathbf{1}\{f(x_{i})=y_{i}\}}{N} $, where $N$ is the total number of the test samples.
    \item \textbf{MAP}: Mean average precision (MAP) is the common evaluation metric for the object detection task, which is the average of AP (average precision) over all object classes.
    \item \textbf{MSE}: Mean Square Error (MSE) measures the $L_{2}$ difference between the predicted output and the ground-truth output after being averaged over the coordinates. Formally, MSE writes $\operatorname{MSE}(f(x),y) =||f(x)-y||_{2}/\operatorname{dim}(y)$, where $\operatorname{dim}(y)$ is the dimension of the ground-truth output. 
\end{itemize}

\subsection{Hiding Single Secret Model}
First, we validate the effectiveness of our proposed  Matryoshka attack in hiding a single secret model in the carrier. Specifically, we choose the secret model from diverse sources while fixing the carrier model as a ResNet-18 on CIFAR-10. We evaluate both the cases when the ParamPool is initialized from the carrier model (marked as \textbf{I}) or initialized with a customized size $5\times$ smaller than the carrier model (marked as \textbf{II}). Table \ref{tab:carrier_suite} reports the performance difference of the carrier and the secret models when comparing with the normal counterparts independently trained (the \textit{Normal} columns). 

\noindent\textbf{Results \& Analysis.} First, from the $\Delta$\textit{Perf}-I column for the carrier model in Table \ref{tab:carrier_suite}, we observe that Matryoshka only sacrifices no more than $1\%$ on the normal utility of the carrier model for covertly transmitting each of the $7$ secret models of diverse nature. According to \cite{gu2017badnet}, a $1\%$ degradation level is indistinguishable to the fluctuation in model performance caused by randomness in training and ensures the attack stealthiness. Meanwhile, from the $\Delta$\textit{Perf}-I column for the secret models, we observe that the performance degradation on the decoded secret models is controlled to be less than $6\%$ uniformly, which indicates a successful functionality stealing \cite{Tramr2016StealingML,CorreiaSilva2018CopycatCS}. Moreover, although the carrier model is normally trained for image classification, we do not observe clear evidence that the carrier model is better in hiding models from the same domain. For example, the $\Delta$\textit{Perf} is slightly larger on the medical imaging dataset DermaMNIST than the audio dataset SpeechCommand (i.e., $-4.78\% < -1.39\%$). According to our analysis in \S\ref{sec:pilot_study}, this may be explained by the observation that an independently optimized M5 model on SpeechCommand has a more similar distribution to the ResNet-18 compared with a LeNet-5 trained on DermaMNIST (i.e., $0.021 < 0.053$ in terms of OTD). Finally, by comparing the $\Delta$\textit{Perf}-I and the $\Delta$\textit{Perf}-II columns in Table \ref{tab:carrier_suite}, we observe the advatange of initializing the ParamPool directly from the carrier model, i.e., Option I, in attack effectiveness. This can be explained by the more learnable parameters provided by the carrier and the more delicate random value scheme adopted by the deep learning framework to initialize the carrier model \cite{Glorot2010UnderstandingTD,He2015DelvingDI}. Therefore, in the following experiments, we mainly evaluate the Option I intialization. Yet, as we have discussed, initialization with a smaller ParamPool, i.e., Option II, has its own advantages in increasing the robustness of Matryoshka, which is later validated in \S\ref{sec:exp:resilience}.  

\subsection{Hiding Multiple Secret Models}
\label{sec:exp:multiple}
Next, we substantially extend the case of single model hiding to covertly transmitting multiple secret models in the carrier model at one time. Specifically, we select a representative set of $6$ secret models in the \textit{Model} column of Table \ref{tab:carrier_multi}, spanning all the $5$ application domains we cover. We leverage the Matryoshka attack with a ParamPool initialized from the carrier model ResNet-18. Table \ref{tab:carrier_multi} reports the performance difference of the carrier and the secret models.

\begin{table}[h]
  \centering
  \caption{Effectiveness of Matryoshka in transmitting multiple secret models (The table excludes $2$ auxiliary models for stealing non-learnable parameters).}
  \scalebox{0.9}{
    \begin{tabular}{cccccc}
    \toprule
          & \textbf{Domain} & \textbf{Model} & \textbf{\# Params}  & $\Delta$\textit{Perf} & \textbf{Normal} \\
    \cmidrule(lr){2-4} \cmidrule(lr){5-6}
    \multirow{7}[4]{*}{\rotatebox[origin=c]{90}{\textbf{Secret Models}}} & \multicolumn{1}{l}{Image} & MobileNetV1-SSD & 9.47M & -1.22\% & 51.57\%	\\\cmidrule(lr){2-6}
          & \multicolumn{1}{l}{Text} & TextCNN & 6.69M & -5.61\% & 90.85\% \\
             \cmidrule(lr){2-6}
          & \multicolumn{1}{l}{{Audio}} & M5 & 26.92K & -1.89\% & 96.86\% \\
              \cmidrule(lr){2-6}
          & \multicolumn{1}{l}{{Time Series}} & RNN & 1.67K & 2.28\% & 86.36\% \\
              \cmidrule(lr){2-6}
          & \multicolumn{1}{l}{\multirow{2}[1]{*}{{Healthcare}}} & LeNet-5 & 61.98K & -6.13\% & 74.21\% \\
          &     & MLP & 0.5K  & -0.27  & 8.23 \\
\cmidrule(lr){2-6}          &        \multicolumn{2}{c}{\textbf{\# Params in Total =}} & \textbf{16.25M} &  -    & - \\
    \midrule
    \midrule
    \multicolumn{2}{c}{\textbf{Carrier Model}} & \textbf{ResNet-18} & 11.17M & -0.41\% & 95.50\% \\
    \bottomrule
    \end{tabular}}%
  \label{tab:carrier_multi}%
\end{table}%


\noindent\textbf{Results \& Analysis.} 
As Table \ref{tab:carrier_multi} shows, Matryoshka also shows high effectiveness in covertly transmitting multiple models at one time. When we encode all the $6$ models and the $2$ auxiliary models in a ResNet-18, the $\Delta$\textit{Perf} on the carrier model remains less than $1\%$ as when encoding a single secret model. This phenomenon is more striking in this case because the carrier model has to encode multiple models  which has at least $30\%$ more parameters than itself (i.e., $11.17$M vs. $16.25$M plus the auxiliary model). Moreover, compared with the results in Table \ref{tab:carrier_suite}, we observe that hiding multiple secret models neither incur more utility losses on the secret models. For example, the $\Delta$\textit{Perf} on TextCNN is $5.61\%$ and $5.65\%$ when it is hidden with or without other $5$ models, while, on RNN, the performance even increases by near $2\%$ when it is hidden with other models. In contrast, most conventional information hiding approaches, including LSB and sign encoding, are unable to hide a larger carrier in a smaller one by design \cite{Provos2003HideAS} and hence cannot attain the same hiding capability as Matryoshka.

\subsection{Exploring the Hiding Capacity}
\begin{figure}[ht]
\begin{center}
\centering
\includegraphics[width=0.45\textwidth]{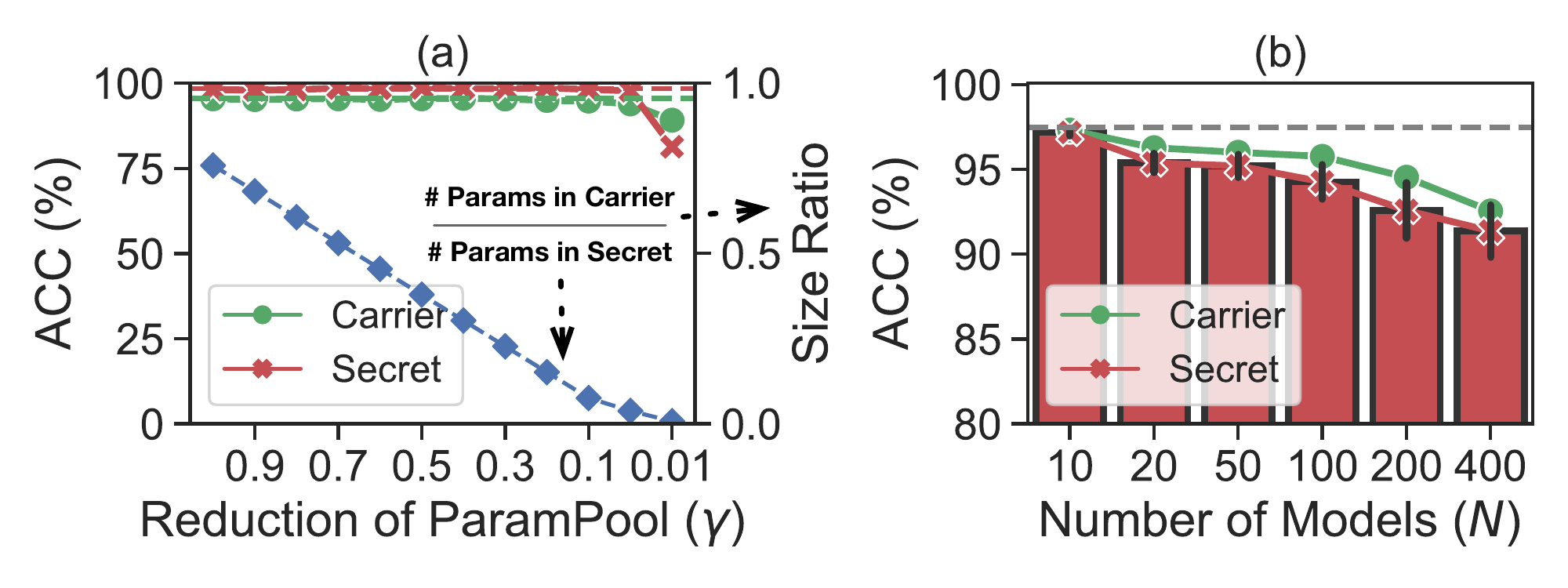}

\caption{The effectiveness of Matryoshka (a) when the ratio between the carrier model and the secret model decreases and (b) when the number of secret models increases. The dashed horizontal lines report the performance of the normal counterparts.}

\label{fig:capacity_limit}
\end{center}
\end{figure}

\noindent\textbf{On the Size of the Secret Model.} To explore the capacity limit of Matryoshka, we further validate the effectiveness in hiding a secret model typically larger than the carrier model by multiple times. Specifically, we initialize the ParamPool with a $\gamma$ (i.e., $\gamma \in (0, 1.0]$) proportion of parameters from the full carrier mode, i.e., a ResNet-18 on CIFAR-10 \cite{krizhevsky2009learning}. We choose the secret model as a VGG-16 on GTSRB \cite{Houben-IJCNN-2013}, which contains $14.74$M parameters, $1.32\times$ of the full carrier model. Then we vary the proportion $\gamma$ from $1.0$ to $0.01$ and conduct Matryoshka with the ParamPool of different sizes. Fig.\ref{fig:capacity_limit}(a) reports the accuracy curves of the secret and the carrier model, where the blue curve reports the size ratio between the secret and the carrier model, and the dashed horizontal lines report the accuracy of the normal counterparts.  

\noindent\textbf{Results \& Analysis.} As Fig.\ref{fig:capacity_limit}(a) shows, Matryoshka is very effective in hiding a large secret model with a much smaller number of parameters from the carrier. For example, when the size of the ParamPool is reduced to $5\%$ of the full size, the carrier model is over $26\times$ smaller than the carrier in terms of the parameter number. Even under this situation,  the accuracy of the secret model and the carrier model remains in a $1\%$ margin of the normal performance. 

From our perspective, Matryoshka's high capacity should be attributed to our novel parameter sharing mechanism, which sufficiently exploits the distribution similarity between model parameters: When a smaller subset of parameters are randomly sampled from the full carrier model, it can still approximate the distribution of the secret model with a tolerable error. As is expected, when the ParamPool size is further reduced to $1\%$ (i.e., $131\times$ smaller), the distortion caused by model hiding is enlarged, i.e., the accuracy of the carrier and the secret models decreases resp. by about $5\%$ and $15\%$.

\noindent\textbf{On the Number of the Secret Models.} We also present the following experiment to explore Matryoshka's behavior when the number of secret tasks $N$ increases. To let $N$ have a near-linear relation to the required hiding capacity, we construct each secret task as a four-layer FCN with the hidden layer sizes randomly sampled from $[100, 300]$ on a permutated MNIST \cite{Goodfellow2014AnEI} with pixels in each input are permuted with a fixed random permutation specific to one task. This ensures each secret model is independent from one another \cite{Cheung2019SuperpositionOM}. The carrier model is an FCN of the average scale, i.e.,  ($784$-$200$-$200$-$10$), on its own permuted MNIST. Fig.\ref{fig:capacity_limit}(b) reports the curve of the average performance of the secret models when $N$ scales from $10$ to $400$, where the dashed line reports the accuracy of an independently trained model which shares the architecture of the carrier model, and the error bars report the standard deviation of the performance of all $N$ secret models. 

\noindent\textbf{Results \& Analysis.} As Fig.\ref{fig:capacity_limit}(b) shows, initially the FCN carrier model encodes $10$ models of the same scale with almost no performance degradation. As the number of tasks continually increases, the accuracy of the carrier model and the secret model decreases only by $1.5\%$ when $N$ scales to $100$, and remains over $90\%$ when $N$ reaches $400$. Combined with the results in \S\ref{sec:exp:multiple}, Matryoshka does allow a carrier model to encode multiple secret models of \textit{different architectures and for different tasks}, with only slight degradation on the normal utility. To the best of knowledge, previous works only explore the superposition of multiple models of the same architecture in one model in the area of continual learning \cite{Cheung2019SuperpositionOM}.

\subsection{Robustness against Popular Post-Processing Techniques}
\label{sec:exp:resilience}
Finally, we evaluate the robustness of Matryoshka when the carrier model undergoes popular model post-processing techniques, including parameter pruning and model fine-tuning. Specifically, we consider the following settings. 

\noindent\textbf{(1) Weight Pruning}: A $\beta$ proportion of weight parameters with the smallest absolute values are set to be $0$ in each layer \cite{Han2015LearningBW}. In our experiments, we vary $\beta$ from $0.1$ to $0.5$.

\noindent\textbf{(2) Partial Finetuning}: To finetune the last $K$ layers of an $H$-layer carrier model, the parameters of the first $H-K$ layers are fixed, while the parameters of the last $K$ layers are further updated by optimizing the learning objective of the carrier model on the training data.

\begin{figure}[ht]
\begin{center}
\includegraphics[width=0.45\textwidth]{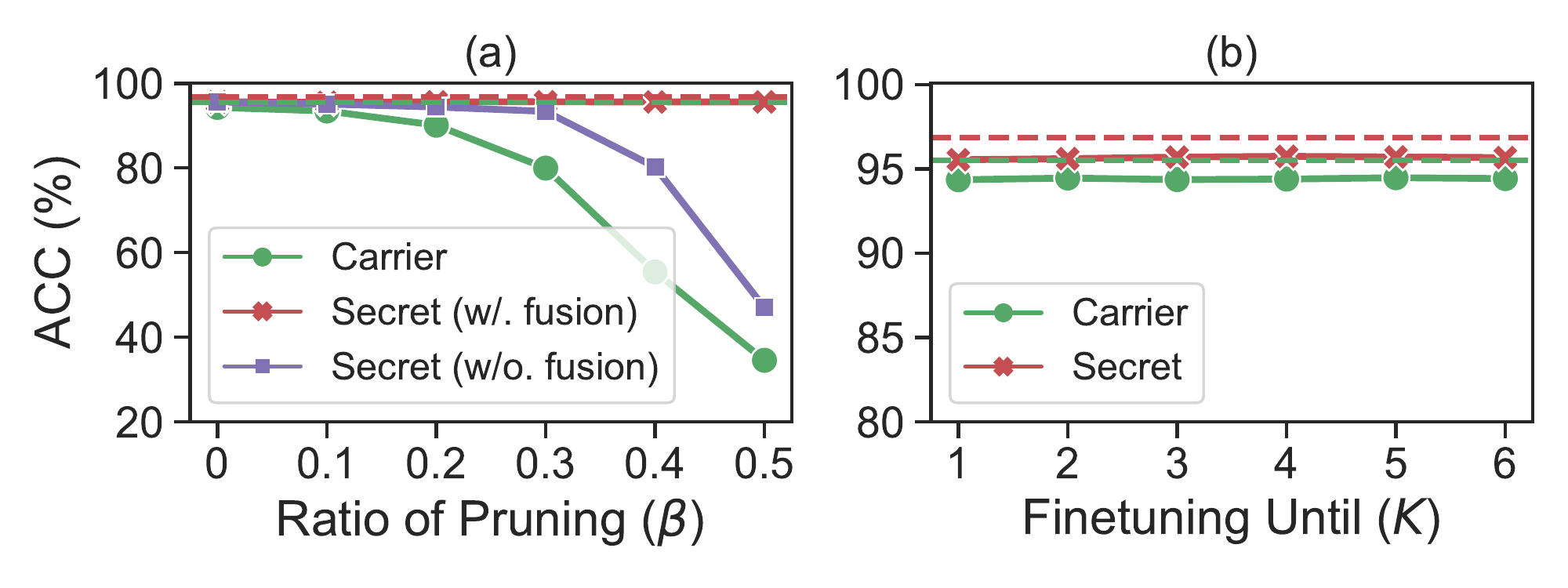}
\vspace{-0.2in}
\caption{Performance of the carrier/secret models under different levels of (a) weight pruning and (b) model finetuning on SpeechCommand. The dashed horizontal lines report the performance of the normal counterparts.}
\label{fig:defense}
\end{center}
\end{figure}

In our experiments, we choose ResNet-18 on CIFAR-10 as the carrier model, with the three models in Table \ref{tab:carrier_suite} on GTSRB (image), IMDB (text) and SpeechCommand (audio) as the secret model respectively. For model fine-tuning, we fine-tune the parameters until the last linear layer, the four basic blocks, and the first convolutional layer in the carrier model respectively. To test fusion techniques at the decoding stage, we initialize the ParamPool with a customized size $5\times$ smaller than the carrier model. This allows us to implement a ParamPool restoration algorithm to recover the pruned parameter by selecting the non-zero value from each ParamPool copy. For fine-tuning, as the first copy of the ParamPool would only be modified when the model is fine-tuned to the first several layers, we always decode the first copy as the ParamPool. Fig.\ref{fig:defense} reports the performance of the carrier and the secret models under different configurations on SpeechCommand. 

\begin{figure}[t]
\begin{center}
\includegraphics[width=0.45\textwidth]{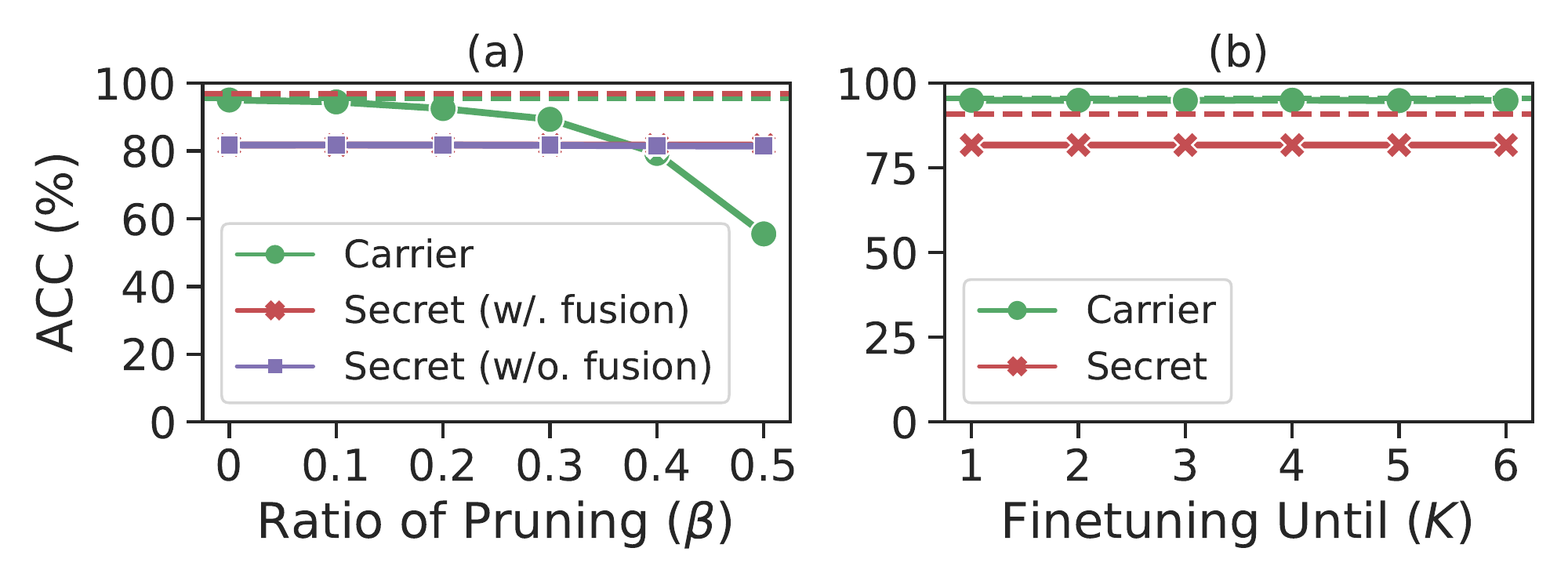}
\caption{Performance of the carrier/secret models under different levels of (a) weight pruning and (b) model finetuning on IMDB.}
\label{fig:defense_textcnn}
\end{center}
\end{figure}

\begin{figure}[t]
\begin{center}
\includegraphics[width=0.45\textwidth]{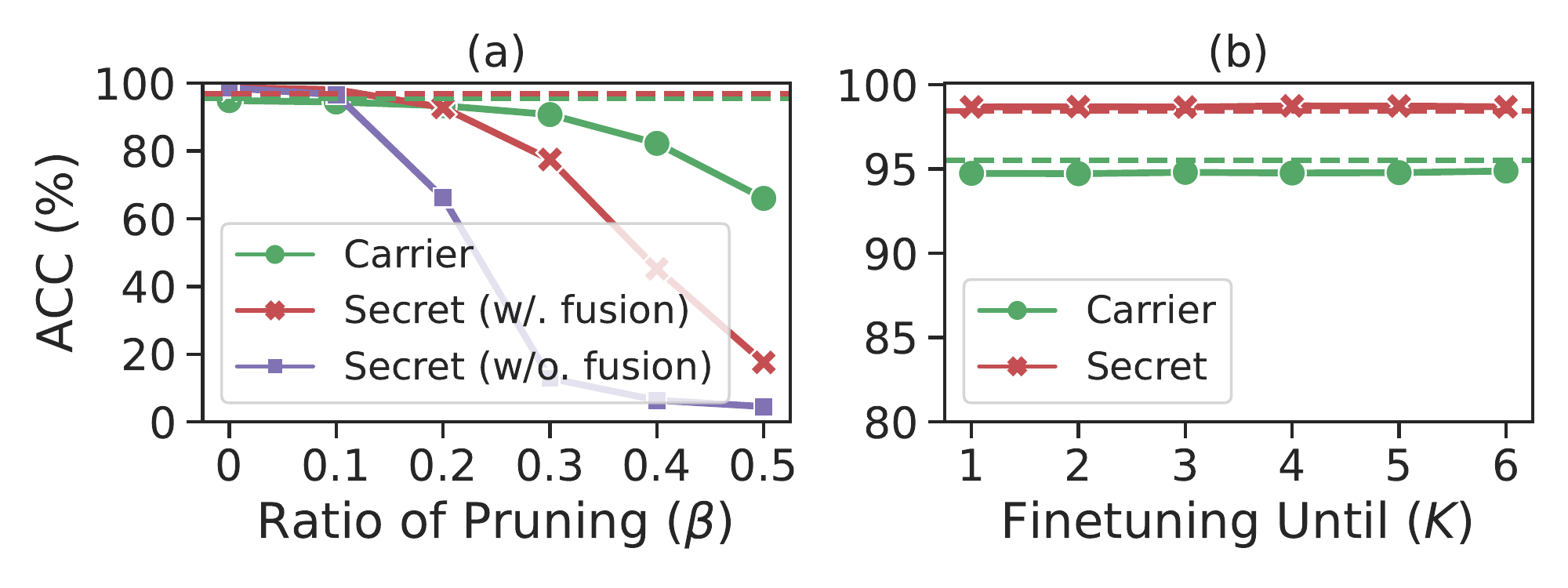}
\caption{Performance of the carrier/secret models under different levels of (a) weight pruning and (b) model finetuning on GTSRB.}
\label{fig:defense_vgg}
\end{center}
\vspace{-0.2in}
\end{figure}

\noindent\textbf{Results \& Analysis.} As Fig.\ref{fig:defense}(a) shows, the performance of the secret model remains high even until the carrier model goes through a huge utility loss due to the large weight pruning ratio. For example, when $\beta$ reaches $0.3$, the performance of the carrier model decreases by over $10\%$ compared with the normal performance (marked in the green dashed line), while the accuracy of the secret model stays within the $1\%$ margin of the normal utility. Comparing the performance of the secret model with and without the fusion smechanisms, we conclude that the robustness of Matryoshka largely comes from the information redunduncy which is innately implemented in our design of ParamPool. Consequently, only if a ParamPool parameter is not always in the $\beta$ smallest for all the layers, our decoding algorithm can always recover its value by referring to the unpruned copy. Similarly from Fig.\ref{fig:defense}(b), robustness is also observed when the carrier model is finetuned until the first layer. Fig.\ref{fig:defense_textcnn}\&\ref{fig:defense_vgg} present similar experimental phenomenon when hiding secret models on texts (IMDB) and images (GTSRB).

Besides the above post-processing techniques we cover, we admit that Matryoshka may be less robust when the architecture of the carrier model is modified. However, it is not a common practice of post-processing an already optimized model by modifying its architecture, except for some special cases that the corporation wants to further optimize the storage and computing efficiency by model compression (e.g., pruning or quantization). Nevertheless, even with model compression, the compressed version is usually released together with the raw full-size model on most third-party platforms \cite{torch_hub,paddle_models}.

Moreover, post-processing is usually conducted by the users after he/she downloads the model and would hardly be conducted by the corporation itself in case of degrading the model utility. In other words, during typical model publishing procedures, the model parameters stay frozen once the training process finishes, so as the architecture of the carrier model. Therefore, the colluder is very likely to obtain the carrier model as in the original form prepared by the insider.   
\subsection{Evading Potential Detection Strategies}
\label{sec:evasion}
To further understand the covertness of our Matryoshka attack, we discuss several potential detection strategies and show our covert transmission strategy can evade them naturally or with slightly more adaptive designs.
Following the same security model in steganography \cite{Provos2003HideAS}, we mainly consider a detector who is able to inspect the published model as a white box to determine whether the model is a carrier model. The detection is assumed to happen before the publishing process. If the detector is confident of the existence of encoded secret models, the transmission would be discarded. However, as a trade-off, when an innocent model is wrongly recognized, the publishing process would be unexpectedly delayed.   

\noindent$\bullet$\textbf{ Detection by Parameter Distribution.} First, we discuss statistics-based steganalysis, which is a classical strategy on multimedia contents \cite{Provos2003HideAS} and may be possibly adopted by a detector who has no knowledge of the attack mechanism. Specifically, the detector compares the model $C$ under inspection with a set of normal models in terms of the parameter distribution to determine whether there is abnormality. In our preliminary study, we mainly consider the situation that the detector has a set of models which are irrelevant with $C$.  We do not consider the case that the detector has a large set of independently trained models of the same architecture as $C$ for doing statistical testing or learning-based detection \cite{Xu2021DetectingAT}. From our viewpoint, this would rarely happen in real world as the purpose of training $C$ is for publishing, which means such a model has not been trained or been available previously. 

\begin{figure}[t]
\begin{center}
\includegraphics[width=0.45\textwidth]{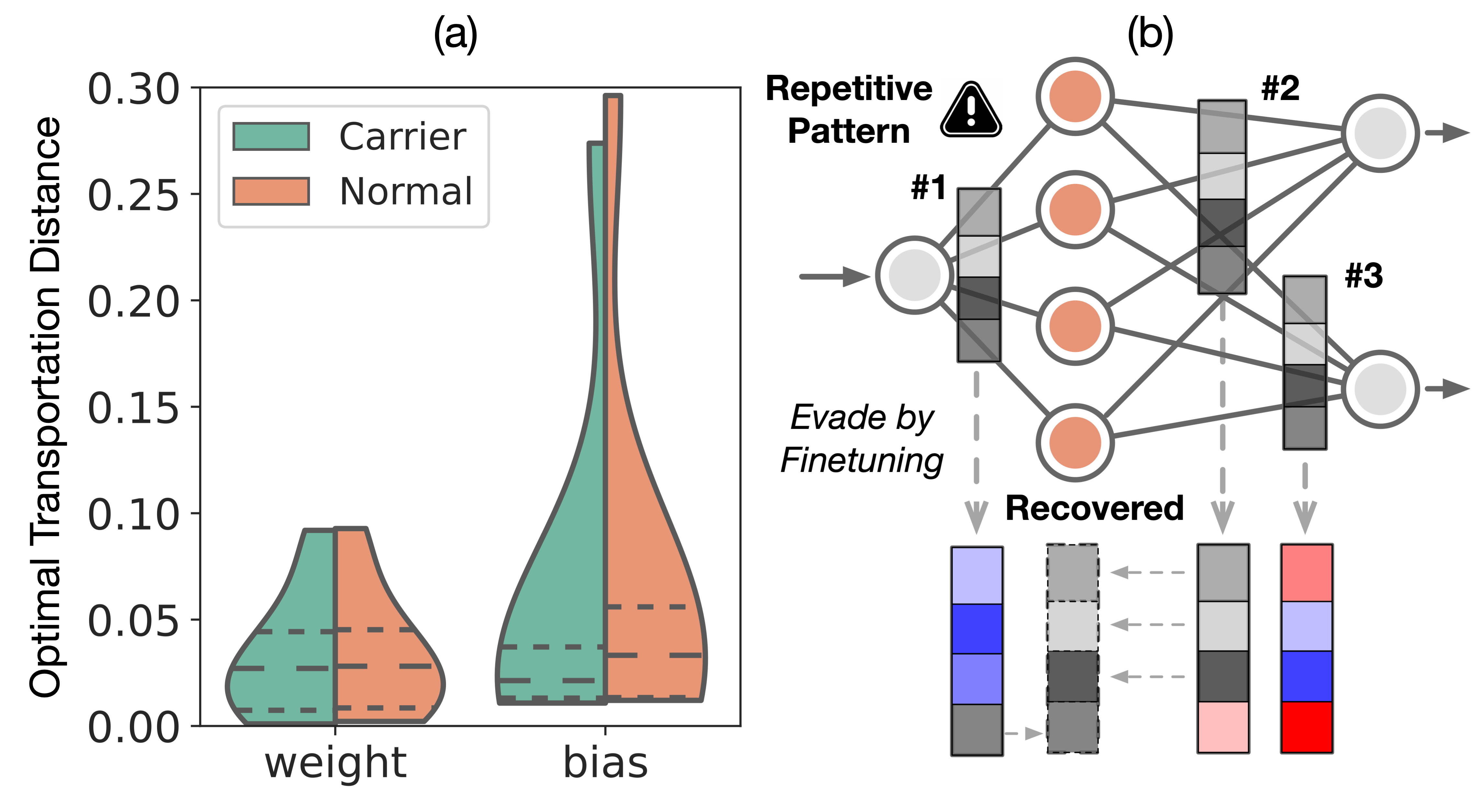}
\caption{(a) Comparison of distributions of parameter dissimilarity, where the dashed lines mark the quantiles. (b) A partial fine-tuning strategy to evade repetitive pattern based detection.}
\label{fig:detection}
\vspace{-0.3in}
\end{center}
\end{figure}

 Specifically, we first prepare a carrier model and a normal model of the same architecture. Then, we calculate the dissimilarity between them with $8$ irrelevant models which we use in \S\ref{sec:pilot_study}. We use the OTD between the parameter distributions to measure the model dissimilarity. Fig.\ref{fig:detection}(a) compares the distribution of the OTD on two types of model pairs, namely, (\textit{carrier}, \textit{irrelevant}) and (\textit{normal}, \textit{irrelevant}), where the distributions of dissimilarity metrics are almost indistinguishable from one another on either weights or biases.

\noindent$\bullet$\textbf{ Detection by Repetitive Patterns.} A more knowledgeable detector who knows the construction procedure of the ParamPool may attempt to analyze the patterns in model parameters for detection. According to \S\ref{sec:attack_pipeline}, when the ParamPool is directly constructed from the carrier model, there should be no abnormal pattern in the carrier model, as each parameter is independent from one another. Besides, according to our previous results, the parameter distribution in the carrier model is also indistinguishable from a normal model. However, an exploitable pattern does exist when the ParamPool is constructed from scratch and has a \textit{smaller} size than the carrier model. As Fig.\ref{fig:detection}(b) shows, a sequence of parameters would occur for multiple times when we collect the learnable parameters in the carrier model sequentially. The detector can exploit this property to run a repetitive subsequence identification algorithm for detection, which has a complexity of $O(n^2)$ (with $n$ the number of parameters in the carrier model). 

Yet, to evade this heuristic-based detection is also straightforward. As Fig.\ref{fig:detection}(b) shows, when the model hiding process terminates, the attacker can actively finetune the carrier model on the training dataset while freezing one copy of ParamPool to eliminate the repetitive parameter patterns. After the active finetuning, neither the carrier model nor the secret models is expected to exhibit performance degradation by design. In our viewpoint, the only cost of this evasion strategy is, after the finetuning, only one copy of ParamPool exists, and no information redundancy can be used to improve the resilience of the secret in the carrier model. We recommend the attacker to determine whether to apply active finetuning or not based on the priority of covertness and robustness.              

\noindent$\bullet$\textbf{ Detection by Monitoring System Usages.} Finally, we discuss the case that the detector monitors the real-time system information of the computing devices to determine whether there is an undergoing Matryoshka attack, which would be typically conducted by a system administrator who is in charge of the computing cluster. Under this situation, the detector may observe side-channel information such as the CPU, GPU and memory loads to recognize any undergoing attack (e.g., detecting cryptojacking \cite{Ning2019CapJackCI}). Although the joint training algorithm of Matryoshka does exhibit a simultaneously high utilization rate of the three types of computing resources, such a pattern of resource utilization is very common for deep learning projects. For example, using multiprocessing when streaming the training batches would incur both high CPU and memory use, while using data parallelism for accelerating the training of a DNN model would cause high utilization rates on multiple GPUs. In other words, a detector can hardly differentiate between an undergoing Matryoshka and a resource-intensive training task from the real-time system usage statistics only. 

In summary, the detector in these cases would face a huge risk of classifying an innocent model wrongly as a carrier model and would therefore absent from alarming and discarding any scheduled-to-publish model. In consideration of the practical threats imposed by our attacks on private datasets even with no exposed interface, we strongly encourage the study of effective defense mechanisms as a future work.    








\section{Conclusion}
In this paper, we design a novel insider attack called Matryoshka which for the first time reveals the possibility of breaking the privacy of ML data even with no exposed interface. Via our novel attack technique of hiding models in model, Matryoshka successfully steals both the functionality aof the private dataset(s) only if the attacker is authorized to access the target dataset in the local network. We provide extensive evaluation results to show Matryoshka is efficient, robust and covert in transmitting the secret models via the scheduled-to-published DNN and exhibits a much higher hiding capacity than known approaches. We hope our discovered vulnerability would alarm AI corporations of the potential risks of any unnecessary access to private datasets even if they are safely stored in the local network.   

\clearpage

\bibliographystyle{IEEEtran}
\bibliography{ref}

\begin{thebibliography}{10}
\providecommand{\url}[1]{#1}
\csname url@samestyle\endcsname
\providecommand{\newblock}{\relax}
\providecommand{\bibinfo}[2]{#2}
\providecommand{\BIBentrySTDinterwordspacing}{\spaceskip=0pt\relax}
\providecommand{\BIBentryALTinterwordstretchfactor}{4}
\providecommand{\BIBentryALTinterwordspacing}{\spaceskip=\fontdimen2\font plus
\BIBentryALTinterwordstretchfactor\fontdimen3\font minus
  \fontdimen4\font\relax}
\providecommand{\BIBforeignlanguage}[2]{{%
\expandafter\ifx\csname l@#1\endcsname\relax
\typeout{** WARNING: IEEEtran.bst: No hyphenation pattern has been}%
\typeout{** loaded for the language `#1'. Using the pattern for}%
\typeout{** the default language instead.}%
\else
\language=\csname l@#1\endcsname
\fi
#2}}
\providecommand{\BIBdecl}{\relax}
\BIBdecl

\bibitem{Bengio2021DeepLF}
Y.~Bengio, Y.~LeCun, and G.~E. Hinton, ``Deep learning for ai,''
  \emph{Communications of the ACM}, vol.~64, pp. 58 -- 65, 2021.

\bibitem{hai_report}
``Building a national ai research resource: A blueprint for the national
  research cloud,''
  \url{https://hai.stanford.edu/sites/default/files/2021-10/HAI_NRCR_2021_0.pdf}.

\bibitem{Wenskay1990IntellectualPP}
D.~L. Wenskay, ``Intellectual property protection for neural networks,''
  \emph{Neural Networks}, 1990.

\bibitem{google_security}
``{Google’s Approach to IT Security},''
  \url{https://static.googleusercontent.com/media/1.9.22.221/en//enterprise/pdf/whygoogle/google-common-security-whitepaper.pdf}.

\bibitem{Tramr2016StealingML}
F.~Tram{\`e}r, F.~Zhang, A.~Juels, M.~K. Reiter, and T.~Ristenpart, ``Stealing
  machine learning models via prediction apis,'' in \emph{USENIX Security
  Symposium}, 2016.

\bibitem{CorreiaSilva2018CopycatCS}
J.~R. Correia-Silva, R.~Berriel, C.~S. Badue, A.~F. de~Souza, and
  T.~Oliveira-Santos, ``Copycat cnn: Stealing knowledge by persuading
  confession with random non-labeled data,'' \emph{2018 International Joint
  Conference on Neural Networks (IJCNN)}, pp. 1--8, 2018.

\bibitem{Chandrasekaran2020ExploringCB}
V.~Chandrasekaran, K.~Chaudhuri, I.~Giacomelli, S.~Jha, and S.~Yan, ``Exploring
  connections between active learning and model extraction,'' in \emph{USENIX
  Security Symposium}, 2020.

\bibitem{Jagielski2020HighAA}
M.~Jagielski, N.~Carlini, D.~Berthelot, A.~Kurakin, and N.~Papernot, ``High
  accuracy and high fidelity extraction of neural networks,'' in \emph{USENIX
  Security Symposium}, 2020.

\bibitem{Hu2021StealingML}
H.~Hu and J.~Pang, ``Stealing machine learning models: Attacks and
  countermeasures for generative adversarial networks,'' \emph{Annual Computer
  Security Applications Conference}, 2021.

\bibitem{Fredrikson2015ModelIA}
M.~Fredrikson, S.~Jha, and T.~Ristenpart, ``Model inversion attacks that
  exploit confidence information and basic countermeasures,'' in \emph{CCS},
  2015.

\bibitem{Ganju2018PropertyIA}
K.~Ganju, Q.~Wang, W.~Yang, C.~A. Gunter, and N.~Borisov, ``Property inference
  attacks on fully connected neural networks using permutation invariant
  representations,'' in \emph{CCS}, 2018.

\bibitem{Melis2019ExploitingUF}
L.~Melis, C.~Song, E.~D. Cristofaro, and V.~Shmatikov, ``Exploiting unintended
  feature leakage in collaborative learning,'' \emph{2019 IEEE Symposium on
  Security and Privacy (SP)}, pp. 691--706, 2019.

\bibitem{Fredrikson2014PrivacyIP}
M.~Fredrikson, E.~Lantz, S.~Jha, S.~Lin, D.~Page, and T.~Ristenpart, ``Privacy
  in pharmacogenetics: An end-to-end case study of personalized warfarin
  dosing,'' vol. 2014, pp. 17--32, 2014.

\bibitem{Shokri2016MembershipIA}
R.~Shokri, M.~Stronati, C.~Song, and V.~Shmatikov, ``Membership inference
  attacks against machine learning models,'' \emph{Security \& Privacy}, pp.
  3--18, 2017.

\bibitem{Salem2019MLLeaksMA}
A.~Salem, Y.~Zhang, M.~Humbert, M.~Fritz, and M.~Backes, ``Ml-leaks: Model and
  data independent membership inference attacks and defenses on machine
  learning models,'' \emph{NDSS}, 2019.

\bibitem{Cheddad2010DigitalIS}
A.~Cheddad, J.~Condell, K.~Curran, and P.~M. Kevitt, ``Digital image
  steganography: Survey and analysis of current methods,'' \emph{Signal
  Process.}, vol.~90, pp. 727--752, 2010.

\bibitem{Yang2019RNNStegaLS}
Z.~Yang, X.~Guo, Z.-M. Chen, Y.~Huang, and Y.-J. Zhang, ``Rnn-stega: Linguistic
  steganography based on recurrent neural networks,'' \emph{IEEE Transactions
  on Information Forensics and Security}, vol.~14, pp. 1280--1295, 2019.

\bibitem{Djebbar2012ComparativeSO}
F.~Djebbar and B.~Ayad, ``Comparative study of digital audio steganography
  techniques,'' \emph{EURASIP Journal on Audio, Speech, and Music Processing},
  vol. 2012, pp. 1--16, 2012.

\bibitem{Song2017MachineLM}
C.~Song, T.~Ristenpart, and V.~Shmatikov, ``Machine learning models that
  remember too much,'' \emph{Proceedings of the 2017 ACM SIGSAC Conference on
  Computer and Communications Security}, 2017.

\bibitem{Uchida2017EmbeddingWI}
Y.~Uchida, Y.~Nagai, S.~Sakazawa, and S.~Satoh, ``Embedding watermarks into
  deep neural networks,'' \emph{Proceedings of the 2017 ACM on International
  Conference on Multimedia Retrieval}, 2017.

\bibitem{Provos2003HideAS}
N.~Provos and P.~Honeyman, ``Hide and seek: An introduction to steganography,''
  \emph{IEEE Secur. Priv.}, vol.~1, pp. 32--44, 2003.

\bibitem{Neumann1956ProbabilisticLA}
J.~von Neumann, ``Probabilistic logic and the synthesis of reliable organisms
  from unreliable components,'' \emph{Automata studies}, vol.~34, pp. 43--98,
  1956.

\bibitem{Ateniese2015HackingSM}
G.~Ateniese, L.~V. Mancini, A.~Spognardi, A.~Villani, D.~Vitali, and G.~Felici,
  ``Hacking smart machines with smarter ones: How to extract meaningful data
  from machine learning classifiers,'' \emph{Int. J. Secur. Networks}, vol.~10,
  pp. 137--150, 2015.

\bibitem{Orekondy2019KnockoffNS}
T.~Orekondy, B.~Schiele, and M.~Fritz, ``Knockoff nets: Stealing functionality
  of black-box models,'' \emph{2019 IEEE/CVF Conference on Computer Vision and
  Pattern Recognition (CVPR)}, pp. 4949--4958, 2019.

\bibitem{Oh2018TowardsRB}
S.~J. Oh, M.~Augustin, M.~Fritz, and B.~Schiele, ``Towards reverse-engineering
  black-box neural networks,'' in \emph{ICLR}, 2018.

\bibitem{Carlini2020CryptanalyticEO}
N.~Carlini, M.~Jagielski, and I.~Mironov, ``Cryptanalytic extraction of neural
  network models,'' in \emph{CRYPTO}, 2020.

\bibitem{Li2021MembershipLI}
Z.~Li and Y.~Zhang, ``Membership leakage in label-only exposures,''
  \emph{Proceedings of the 2021 ACM SIGSAC Conference on Computer and
  Communications Security}, 2021.

\bibitem{ChoquetteChoo2021LabelOnlyMI}
C.~A. Choquette-Choo, F.~Tram{\`e}r, N.~Carlini, and N.~Papernot, ``Label-only
  membership inference attacks,'' in \emph{ICML}, 2021.

\bibitem{Pan2020PrivacyRO}
X.~Pan, M.~Zhang, S.~Ji, and M.~Yang, ``Privacy risks of general-purpose
  language models,'' \emph{2020 IEEE Symposium on Security and Privacy (SP)},
  pp. 1314--1331, 2020.

\bibitem{Zhu2019DeepLF}
L.~Zhu, Z.~Liu, and S.~Han, ``Deep leakage from gradients,'' in \emph{NeurIPS},
  2019.

\bibitem{Geiping2020InvertingG}
J.~Geiping, H.~Bauermeister, H.~Dr{\"o}ge, and M.~Moeller, ``Inverting
  gradients - how easy is it to break privacy in federated learning?'' in
  \emph{NeurIPS}, 2020.

\bibitem{Carlini2019TheSS}
N.~Carlini, C.~Liu, {\'U}.~Erlingsson, J.~Kos, and D.~X. Song, ``The secret
  sharer: Evaluating and testing unintended memorization in neural networks,''
  in \emph{USENIX Security Symposium}, 2019.

\bibitem{Salem2020UpdatesLeakDS}
A.~Salem, A.~Bhattacharyya, M.~Backes, M.~Fritz, and Y.~Zhang, ``Updates-leak:
  Data set inference and reconstruction attacks in online learning,''
  \emph{USENIX Security}, 2020.

\bibitem{Radhakrishnan2020OverparameterizedNN}
A.~Radhakrishnan, M.~Belkin, and C.~Uhler, ``Overparameterized neural networks
  implement associative memory,'' \emph{Proceedings of the National Academy of
  Sciences of the United States of America}, vol. 117, pp. 27\,162 -- 27\,170,
  2020.

\bibitem{Carlini2021ExtractingTD}
N.~Carlini, F.~Tram{\`e}r, E.~Wallace, M.~Jagielski, A.~Herbert-Voss, K.~Lee,
  A.~Roberts, T.~B. Brown, D.~X. Song, {\'U}.~Erlingsson, A.~Oprea, and
  C.~Raffel, ``Extracting training data from large language models,'' in
  \emph{USENIX Security Symposium}, 2021.

\bibitem{Adi2018TurningYW}
Y.~Adi, C.~Baum, M.~Ciss{\'e}, B.~Pinkas, and J.~Keshet, ``Turning your
  weakness into a strength: Watermarking deep neural networks by backdooring,''
  in \emph{USENIX Security Symposium}, 2018.

\bibitem{Chen2019DeepMarksAS}
H.~Chen, B.~D. Rouhani, C.~Fu, J.~Zhao, and F.~Koushanfar, ``Deepmarks: A
  secure fingerprinting framework for digital rights management of deep
  learning models,'' \emph{Proceedings of the 2019 on International Conference
  on Multimedia Retrieval}, 2019.

\bibitem{Wang2021RIGACA}
T.~Wang and F.~Kerschbaum, ``Riga: Covert and robust white-box watermarking of
  deep neural networks,'' \emph{Proceedings of the Web Conference 2021}, 2021.

\bibitem{Goodfellow-et-al-2016}
I.~Goodfellow, Y.~Bengio, and A.~Courville, \emph{Deep Learning}.\hskip 1em
  plus 0.5em minus 0.4em\relax MIT Press, 2016.

\bibitem{Heaton2016DeepLF}
J.~B. Heaton, N.~G. Polson, and J.~H. Witte, ``Deep learning for finance: Deep
  portfolios,'' \emph{Econometric Modeling: Capital Markets - Portfolio Theory
  eJournal}, 2016.

\bibitem{Esteva2017DermatologistlevelCO}
A.~Esteva, B.~Kuprel, R.~A. Novoa, J.~M. Ko, S.~M. Swetter, H.~M. Blau, and
  S.~Thrun, ``Dermatologist-level classification of skin cancer with deep
  neural networks,'' \emph{Nature}, 2017.

\bibitem{Redmon2016YouOL}
J.~Redmon, S.~K. Divvala, R.~B. Girshick, and A.~Farhadi, ``You only look once:
  Unified, real-time object detection,'' \emph{2016 IEEE Conference on Computer
  Vision and Pattern Recognition (CVPR)}, pp. 779--788, 2016.

\bibitem{Goodfellow2015DeepL}
I.~J. Goodfellow, Y.~Bengio, and A.~C. Courville, ``Deep learning,''
  \emph{Nature}, vol. 521, pp. 436--444, 2015.

\bibitem{Kingma2015AdamAM}
D.~P. Kingma and J.~Ba, ``Adam: A method for stochastic optimization,''
  \emph{CoRR}, vol. abs/1412.6980, 2015.

\bibitem{Alneyadi2016ASO}
S.~Alneyadi, E.~Sithirasenan, and V.~Muthukkumarasamy, ``A survey on data
  leakage prevention systems,'' \emph{J. Netw. Comput. Appl.}, vol.~62, pp.
  137--152, 2016.

\bibitem{securosis_report}
``{Understanding and Selecting a Data Loss Prevention Solution},''
  \url{https://securosis.com/assets/library/reports/DLP-Whitepaper.pdf}.

\bibitem{Ren2015FasterRT}
S.~Ren, K.~He, R.~B. Girshick, and J.~Sun, ``Faster r-cnn: Towards real-time
  object detection with region proposal networks,'' \emph{IEEE Transactions on
  Pattern Analysis and Machine Intelligence}, vol.~39, pp. 1137--1149, 2015.

\bibitem{Shorten2019ASO}
C.~Shorten and T.~M. Khoshgoftaar, ``A survey on image data augmentation for
  deep learning,'' \emph{Journal of Big Data}, vol.~6, pp. 1--48, 2019.

\bibitem{Baltruaitis2019MultimodalML}
T.~Baltruaitis, C.~Ahuja, and L.-P. Morency, ``Multimodal machine learning: A
  survey and taxonomy,'' \emph{IEEE Transactions on Pattern Analysis and
  Machine Intelligence}, vol.~41, pp. 423--443, 2019.

\bibitem{Krombholz2015AdvancedSE}
K.~Krombholz, H.~Hobel, M.~Huber, and E.~R. Weippl, ``Advanced social
  engineering attacks,'' \emph{J. Inf. Secur. Appl.}, vol.~22, pp. 113--122,
  2015.

\bibitem{google_bert}
``{Open Sourcing BERT},''
  \url{https://ai.googleblog.com/2018/11/open-sourcing-bert-state-of-art-pre.html}.

\bibitem{openai_git}
``{GitHub Homepage of OpenAI},'' \url{https://github.com/openai}.

\bibitem{Kurak1992ACN}
C.~W. Kurak and J.~McHugh, ``A cautionary note on image downgrading,''
  \emph{[1992] Proceedings Eighth Annual Computer Security Application
  Conference}, pp. 153--159, 1992.

\bibitem{Liu2020StegoNetTD}
T.~Liu, Z.~Liu, Q.~Liu, W.~Wen, W.~Xu, and M.~Li, ``Stegonet: Turn deep neural
  network into a stegomalware,'' \emph{Annual Computer Security Applications
  Conference}, 2020.

\bibitem{bossbase}
``{BOSSBase Stegnoanalysis Dataset},''
  \url{http://agents.fel.cvut.cz/boss/index.php?mode=VIEW&tmpl=materials}.

\bibitem{Jacob2018QuantizationAT}
B.~Jacob, S.~Kligys, B.~Chen, M.~Zhu, M.~Tang, A.~G. Howard, H.~Adam, and
  D.~Kalenichenko, ``Quantization and training of neural networks for efficient
  integer-arithmetic-only inference,'' \emph{2018 IEEE/CVF Conference on
  Computer Vision and Pattern Recognition}, pp. 2704--2713, 2018.

\bibitem{Krishnamoorthi2018QuantizingDC}
R.~Krishnamoorthi, ``Quantizing deep convolutional networks for efficient
  inference: A whitepaper,'' \emph{ArXiv}, vol. abs/1806.08342, 2018.

\bibitem{apple_hub}
``{PyTorch Hub},'' \url{https://pytorch.org/hub/}.

\bibitem{Rubner2004TheEM}
Y.~Rubner, C.~Tomasi, and L.~J. Guibas, ``The earth mover's distance as a
  metric for image retrieval,'' \emph{International Journal of Computer
  Vision}, vol.~40, pp. 99--121, 2004.

\bibitem{villani2008optimal}
C.~Villani, \emph{Optimal transport: old and new}.\hskip 1em plus 0.5em minus
  0.4em\relax Springer Science \& Business Media, 2008, vol. 338.

\bibitem{Weng2020TowardsCM}
T.-W. Weng, P.~Zhao, S.~Liu, P.-Y. Chen, X.~Lin, and L.~Daniel, ``Towards
  certificated model robustness against weight perturbations,'' in \emph{AAAI},
  2020.

\bibitem{Houben-IJCNN-2013}
S.~Houben, J.~Stallkamp, J.~Salmen, M.~Schlipsing, and C.~Igel, ``Detection of
  traffic signs in real-world images: The {G}erman {T}raffic {S}ign {D}etection
  {B}enchmark,'' in \emph{International Joint Conference on Neural Networks},
  no. 1288, 2013.

\bibitem{simonyan2014very}
K.~Simonyan and A.~Zisserman, ``Very deep convolutional networks for
  large-scale image recognition,'' \emph{arXiv preprint arXiv:1409.1556}, 2014.

\bibitem{Everingham15}
M.~Everingham, S.~M.~A. Eslami, L.~Van~Gool, C.~K.~I. Williams, J.~Winn, and
  A.~Zisserman, ``The pascal visual object classes challenge: A
  retrospective,'' \emph{International Journal of Computer Vision}, vol. 111,
  no.~1, pp. 98--136, Jan. 2015.

\bibitem{howard2017mobilenets}
A.~G. Howard, M.~Zhu, B.~Chen, D.~Kalenichenko, W.~Wang, T.~Weyand,
  M.~Andreetto, and H.~Adam, ``Mobilenets: Efficient convolutional neural
  networks for mobile vision applications,'' \emph{arXiv preprint
  arXiv:1704.04861}, 2017.

\bibitem{maas-EtAl:2011:ACL-HLT2011}
\BIBentryALTinterwordspacing
A.~L. Maas, R.~E. Daly, P.~T. Pham, D.~Huang, A.~Y. Ng, and C.~Potts,
  ``Learning word vectors for sentiment analysis,'' in \emph{Proceedings of the
  49th Annual Meeting of the Association for Computational Linguistics: Human
  Language Technologies}.\hskip 1em plus 0.5em minus 0.4em\relax Portland,
  Oregon, USA: Association for Computational Linguistics, June 2011, pp.
  142--150. [Online]. Available: \url{http://www.aclweb.org/anthology/P11-1015}
\BIBentrySTDinterwordspacing

\bibitem{zhang2015sensitivity}
Y.~Zhang and B.~Wallace, ``A sensitivity analysis of (and practitioners' guide
  to) convolutional neural networks for sentence classification,'' \emph{arXiv
  preprint arXiv:1510.03820}, 2015.

\bibitem{speechcommandsv2}
\BIBentryALTinterwordspacing
P.~{Warden}, ``{Speech Commands: A Dataset for Limited-Vocabulary Speech
  Recognition},'' \emph{ArXiv e-prints}, Apr. 2018. [Online]. Available:
  \url{https://arxiv.org/abs/1804.03209}
\BIBentrySTDinterwordspacing

\bibitem{7952190}
W.~Dai, C.~Dai, S.~Qu, J.~Li, and S.~Das, ``Very deep convolutional neural
  networks for raw waveforms,'' in \emph{2017 IEEE International Conference on
  Acoustics, Speech and Signal Processing (ICASSP)}, 2017, pp. 421--425.

\bibitem{liu2009uwave}
J.~Liu, L.~Zhong, J.~Wickramasuriya, and V.~Vasudevan, ``uwave:
  Accelerometer-based personalized gesture recognition and its applications,''
  \emph{Pervasive and Mobile Computing}, vol.~5, no.~6, pp. 657--675, 2009.

\bibitem{medmnistv2}
J.~Yang, R.~Shi, D.~Wei, Z.~Liu, L.~Zhao, B.~Ke, H.~Pfister, and B.~Ni,
  ``Medmnist v2: A large-scale lightweight benchmark for 2d and 3d biomedical
  image classification,'' \emph{arXiv preprint arXiv:2110.14795}, 2021.

\bibitem{lecun1998gradient}
Y.~LeCun, L.~Bottou, Y.~Bengio, and P.~Haffner, ``Gradient-based learning
  applied to document recognition,'' \emph{Proceedings of the IEEE}, vol.~86,
  no.~11, pp. 2278--2324, 1998.

\bibitem{Truda2020warfit}
\BIBentryALTinterwordspacing
G.~Truda and P.~Marais, ``Evaluating warfarin dosing models on multiple
  datasets with a novel software framework and evolutionary optimisation,''
  \emph{Journal of Biomedical Informatics}, p. 103634, 2020. [Online].
  Available:
  \url{http://www.sciencedirect.com/science/article/pii/S1532046420302628}
\BIBentrySTDinterwordspacing

\bibitem{gu2017badnet}
T.~Gu, K.~Liu, B.~Dolan-Gavitt, and S.~Garg, ``Badnets: Evaluating backdooring
  attacks on deep neural networks,'' \emph{IEEE Access}, vol.~7, pp.
  47\,230--47\,244, 2019.

\bibitem{Glorot2010UnderstandingTD}
X.~Glorot and Y.~Bengio, ``Understanding the difficulty of training deep
  feedforward neural networks,'' in \emph{AISTATS}, 2010.

\bibitem{He2015DelvingDI}
K.~He, X.~Zhang, S.~Ren, and J.~Sun, ``Delving deep into rectifiers: Surpassing
  human-level performance on imagenet classification,'' \emph{2015 IEEE
  International Conference on Computer Vision (ICCV)}, pp. 1026--1034, 2015.

\bibitem{krizhevsky2009learning}
A.~Krizhevsky, ``Learning multiple layers of features from tiny images,'' 2009.

\bibitem{Goodfellow2014AnEI}
I.~J. Goodfellow, M.~Mirza, X.~Da, A.~C. Courville, and Y.~Bengio, ``An
  empirical investigation of catastrophic forgeting in gradient-based neural
  networks,'' \emph{ICLR}, 2014.

\bibitem{Cheung2019SuperpositionOM}
B.~Cheung, A.~Terekhov, Y.~Chen, P.~Agrawal, and B.~A. Olshausen,
  ``Superposition of many models into one,'' \emph{ArXiv}, vol. abs/1902.05522,
  2019.

\bibitem{Han2015LearningBW}
S.~Han, J.~Pool \emph{et~al.}, ``Learning both weights and connections for
  efficient neural network,'' \emph{ArXiv}, 2015.

\bibitem{torch_hub}
``{Models - Machine Learning - Apple Developer},''
  \url{https://developer.apple.com/machine-learning/models/}.

\bibitem{paddle_models}
``{Available Models-PaddleLite},''
  https://paddle-lite.readthedocs.io/zh/latest/introduction/support\_model\_list.html,
  accessed: 2021-05-21.

\bibitem{Xu2021DetectingAT}
X.~Xu, Q.~Wang, H.~Li, N.~Borisov, C.~A. Gunter, and B.~Li, ``Detecting ai
  trojans using meta neural analysis,'' \emph{2021 IEEE Symposium on Security
  and Privacy (SP)}, pp. 103--120, 2021.

\bibitem{Ning2019CapJackCI}
R.~Ning, C.~Wang, C.~Xin, J.~Li, L.~Zhu, and H.~Wu, ``Capjack: Capture
  in-browser crypto-jacking by deep capsule network through behavioral
  analysis,'' \emph{IEEE INFOCOM 2019 - IEEE Conference on Computer
  Communications}, pp. 1873--1881, 2019.

\end{thebibliography}

\end{document}